\pdfoutput=1
\documentclass[9pt,shortpaper,twoside,web]{Packages/ieeecolor}

\usepackage{Packages/generic}
\usepackage{cite}
\usepackage{amsmath,amssymb,amsfonts}
\usepackage{graphicx}
\usepackage{textcomp}
\usepackage{import}

\usepackage[printonlyused,withpage,nolist]{acronym}
\usepackage{algorithm,algpseudocode}
\usepackage{bbm}
\usepackage{bm}
\usepackage{booktabs}

\usepackage{xcolor}

\usepackage{multirow}
\usepackage{optidef}
\usepackage[belowskip=-15pt,aboveskip=5pt]{caption}

\usepackage{soul}

\usepackage{tabularx} 
\providecommand{\href}[2]{#2}
\providecommand{\url}[1]{\texttt{#1}}

\graphicspath{{./}{Figures/Framework/SpaceMapping/}{Figures/Framework/ControlFramework/}{Figures/SimResults/BMArm/}{Figures/SimResults/SpiderArm/}{Figures/SimResults/FASTKIT/}{Figures/SimResults/All/}{Figures/HardwareResults/}{Authors_Bio/author/}}

\DeclareMathOperator*{\argmin}{arg\,min}
\newcommand{\norm}[1]{\left\lVert#1\right\rVert}

\newcommand{\citep}{\cite}
\newcommand{\fs}{0.80}
\newcommand{\fss}{0.70}

\newtheorem{theorem}{Theorem}

\newtheorem{remark}{Remark}
\newtheorem{assumption}{Assumption}

\newcommand{\bea}{\begin{IEEEeqnarray}{C}}
	\newcommand{\eea}{\end{IEEEeqnarray}}
\newcommand{\beas}{\begin{IEEEeqnarray*}{C}}
	\newcommand{\eeas}{\end{IEEEeqnarray*}}
\def\btheta{\mbox{\boldmath${\theta}$}}

\def\BibTeX{{\rm B\kern-.05em{\sc i\kern-.025em b}\kern-.08em
		T\kern-.1667em\lower.7ex\hbox{E}\kern-.125emX}}
\begin{document}
	\title{Tri-Space Operational Control of Redundant Multilink and Hybrid Cable-Driven Parallel Robots Using an Iterative-Learning based Reactive Approach}
	\author{
		Dipankar Bhattacharya,~\IEEEmembership{Member,~IEEE,} 
		Yin Pok Chan,  Siqi Shang, Yuen Shan Chan, 
		Ying Tan, ~\IEEEmembership{Fellow,~IEEE,} and 
		Darwin Lau,~\IEEEmembership{Senior Member,~IEEE}
		\thanks{D. Bhattacharya, Y. P. Chan, S. Shang, Y. S. Chan, and D. Lau are with the Department of Mechanical and Automation Engineering, The Chinese University of Hong Kong, Hong Kong SAR.  (email: dipankarbhattacharya@cuhk.edu.hk, ypchan@cuhk.edu.hk, sqshang@link.cuhk.edu.hk, yschan03@cuhk.edu.hk, darwinlau@mae.cuhk.edu.hk.)}
		\thanks{Ying Tan is with the Faculty of Engineering and
			Information Technology, The University of Melbourne, Parkville, VIC 3010,
			Australia (e-mail: yingt@unimelb.edu.au).}
		\thanks{This paper was published in \emph{IEEE Transactions on Control Systems Technology}, vol.~31, no.~6, pp.~2465--2483, Nov.~2023, DOI: 10.1109/TCST.2023.3263689. Licensed under a Creative Commons Attribution 4.0 International License (CC BY 4.0). \textcopyright\ 2023 IEEE.}
	}
	
	\maketitle
	\begin{acronym}[TDMA]
		\acro{ATR}{Avoidance Transition Region}
		\acro{BMArm}{Biologically-inspired Mechanical Arm}
		
		\acro{CASPR}{Cable-Robot Analysis and Simulation Platform for Research}
		\acro{CDPR}{Cable-Driven Parallel Robot}
		
		\acro{DoF}{Degrees of Freedom}
		
		\acro{EoM}{Equation of Motion}
		
		\acro{FD}{Forward Dynamics}
		\acro{FK}{Forward Kinematics}
		
		\acro{HCDR}{Hybrid Cable-Driven Robot}
		
		\acro{ID}{Inverse Dynamics}
		\acro{ILC}{Iterative-Learning Control}
		\acro{IK}{Inverse Kinematics}	
		\acro{IT}{Interference Time}
		
		\acro{MCDR}{Multilink Cable-Driven Robot}
		
		\acro{PS}{Pattern Search}
		\acro{PSO}{Particle Swarm Optimization}
		
		\acro{QP}{Quadratic Programming}
		
		\acro{RC}{Reactive Control}
		\acro{RMSE}{Root Mean Square Error}
		\acro{ROS}{Robot Operating System}
		
		\acro{SCDR}{Single-link Cable-Driven Robot}
	\end{acronym}
	
	\begin{abstract}
		\textit{\acp{CDPR}} are a type of parallel mechanism in which cables are used as actuators. Due to the two levels of redundancy and numerous constraints within the CDPR actuation, joint and operational spaces (together known as the \textit{tri-space}), tracking a given trajectory in the operational space while satisfying constraints in tri-space simultaneously is challenging. To the best of the authors' knowledge, there does not exist any tri-space control framework, which is robust, effective, and directly applicable to several architectures of redundantly actuated CDPRs. This paper proposes a tri-space control framework that combines Reactive Control (RC) and Iterative-Learning Control (ILC) to perform  repetitive tasks in the operational space. The framework allows the tracking of operational space trajectories online with feasible cable forces, while avoiding undesirable situations such as cable-link interference, joint interference, and loss of manipulability. On the other hand, by finding an optimal parameter in the null space using a novel parameterization of a null
		space vector, the performance can be improved through \ac{ILC} when the task is repeatedly executed. Simulation and hardware results on various Multilink Cable-Driven Robot (MCDRs) and Hybrid Cable-Driven Robots(HCDRs) show that the proposed tri-space control framework can be conveniently and effectively applied to the real-time control of different CDPRs.
	\end{abstract}
	
	
	\acresetall
	
	\section{Introduction}
	
	\textit{\acp{CDPR}} are a type of parallel mechanism actuated by cables instead of rigid links, having advantages of low inertia, large potential workspace, high reconfigurability and transportability. Hence, CDPRs have been widely studied and applied in manufacturing \citep{Albus_Bostelman_Dagalakis_93_NIST_RoboCrane, Oh_Mankala_Agrawal_Albus_05_DualStageCargo}, building construction \citep{Bosscher_WilliamsII_Bryson_CastroLacouture_07_ContourCrafting,Wu_Cheng_Fingrut_Crolla_Yam_Lau_18_CUBrick}, rehabilitation \citep{Mao_Agrawal_12_CAREX_Exoskeleton_Rehab}, and rapid prototyping
	\citep{Izard_Dubor_Herve_Cabay_Culla_Rodriguez_Barrado_17_3DPrintCDPR}. However, the control of CDPRs is difficult due to the unique property that cables can only apply pulling forces (\emph{positive cable forces}), resulting in a redundantly actuated system. This redundancy generates challenges in their mechanical design \citep{Lau_Bhalerao_Oetomo_Halgamuge_12_TaskSpecificOptimisation}, workspace analysis \citep{Bouchard_Gosselin_Moore_10_WFW_Prescribed_Wrench, Hassan_Khajepour_11_WCW_Dykstra, Lau_Oetomo_16_WrenchClosureValidityNecessaryConditions, Gouttefarde_Daney_Merlet_11_WFW_Interval_Analysis, Rezazadeh_Behzadipour_11_Multilink_WCW_Analysis}, and control and synthesis \citep{Alp_Agrawal_02_LyapunovFeedbackLinearControl, Fang_Franitza_Torlo_Bekes_Hiller_04_Control_Optimal_Distribution, Oh_Agrawal_05_Control_Positive_Tension, Hassan_Khajepour_08_Positive_Tension_Convex_Analysis}.
	
	\textit{\acp{SCDR}}, a common type of CDPR, consists of a single rigid body end-effector actuated by cables connected from the base frame. However, this design requires the base frame to completely encapsulate the desired workspace, leading to a large robot footprint. Moreover, SCDRs typically have a limited range of end-effector orientations.  \textit{\acp{MCDR}} and \textit{\acp{HCDR}} have been developed to relax such limitations.
	
	MCDRs are a type of CDPR consisting of multiple rigid links in order to combine the dexterity and increased end-effector orientation capability of serial manipulators with the actuation advantages of CDPRs. Additionally, the similarities between MCDRs and anthropomorphic systems have motivated bio-robotic applications, such as musculoskeletal arms \citep{Yang_Lin_Mustafa_Pham_Yeo_05_Multilink_7DOF_Arm},  humanoid \citep{Mustafa_Agrawal_11_Multilink_Min_Cables} robots, and snake-like robots \citep{Racioppo_Tzvi_17_ControlOfSnakeRobot}. HCDRs are hybrid actuated systems that combine cables and direct joint actuation, and possess both passive and active joint \textit{\ac{DoF}}. Examples of HCDRs include: a) the FASTKIT robot \citep{Rasheed_Long_Marquez_Caro_18_FASTKIT}, where a CDPR is mounted onto multiple mobile robot platforms such that these mobile robots can enlarge their workspace and reconfigure the attachment points; and b) the SpiderArm robot, where robot arms are mounted onto the end-effector of an SCDR to increase the dexterity and accuracy of the CDPR while still maintaining a large translational workspace. In this work, CDPRs will be referred to as MCDRs and HCDRs.
	
\begin{figure}[tb!]
		\centering
		\def\svgwidth{1\columnwidth}
		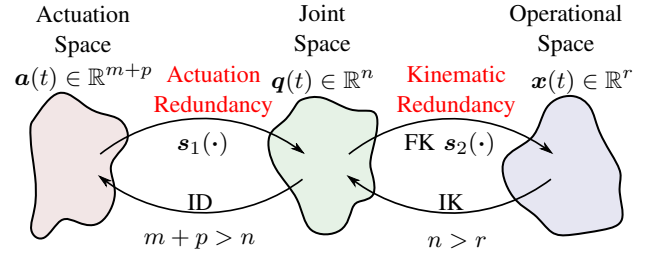
		\caption{Tri-space mapping (shown by arrows) between the actuation, joint and operational spaces of MCDRs and HCDRs for $m+p$ actuators, $n$-dimensonal joint space, and $r$-dimensional operational space such that $ {\bm{s}}_1:\bm{a}\rightarrow \bm{q}  $ and $ {\bm{s}}_2:\bm{q}\rightarrow \bm{r}  $. In this figure ID means Inverse Dynamics, FK means Forward Kinematics and IK means Inverse Kinematics.}
		\label{fig:actuation_joint_operational_spaces}
	\end{figure}

	Compared to SCDRs, the control of CDPRs are more complicated since the kinematics and dynamics have to be described within three spaces: 1) \emph{actuation space} refers to the control input to the robot, such as cable forces and length commands for the actuated \ac{DoF} with $m+p$ DoF; 2) \emph{joint space} represents the DoF (generalized coordinates) with $n$ DoF; and the 3) \emph{operational space} (or \emph{task space}) denotes the pose of the end-effector with the DoF of $r$. The combination if actuation, joint, and operational spaces is defined as the tri-space. The dimensions in the tri-space usually satisfy $ r<n<m+p $ as shown in Fig. \ref{fig:actuation_joint_operational_spaces}). For example, in the SpiderArm robot, $r=6$, $n=12$, and $m+p=8 + 6=14$.
	
	Due to the tri-space setting in CDPRs, the control of CDPRs has two levels of redundancies. One is between operational and joint spaces (\emph{kinematic redundancy}) and the other is between joint and cable actuation spaces (\emph{actuation redundancy}). The kinematic redundancy leads to potentially infinite joint poses that result in the same operational space motion and the actuation redundancy leads to potentially infinite sets of actuation commands to produce the same joint motion. The infinitely many solutions from redundancies always conflict with various constraints, such as positive cable forces, joint limits, and cable interference. 
	
	{Four requirements are identified in developing the tri-space control framework.  Firstly, the framework can solve the tri-space redundancy problem in the CDPRs by simultaneously determining actuation and joint space motion variables in a  single process with the consideration of tri-space constraints.
	{Secondly, the avoidance functions in the framework can guide the CDPRs to smoothly avoid undesirable situations such as cable-link interference, loss of manipulability, and reaching mechanical joint limits in advance. }Thirdly, it can improve the performance if the operational task repeats itself. Lastly, this framework should provide a feasible solution with a relatively lower computational cost.} Consequently, this work proposes a generalized tri-space operational control framework by combining \textit{\ac{RC}} and \textit{\ac{ILC}} for the CDPRs. 

	\textit{Direct reactive operational space control}  or \ac{RC} approaches have been used to directly determine a set of feasible actuation commands at each time step from an operational space trajectory \cite{Khatib_87_UnifiedOpSpaceFormulation}. {However, a drawback for RC is that it only considers the instantaneous pose and does not consider evolution of the pose and control input. Typically, the
	task performance is based on the entire trajectory, where reactive methods struggle.} {Besides, existing} {\ac{RC}} approaches \cite{DeSapio_Warren_Khatib_Delp_05_SimulatingTaskLevelControl} {are unable to handle constraints, such as lack of manipulability} \cite{Jin_2020_Perturbed}, joint limits\cite{Shimizu_2008_Analytical} or cable interference\cite{Lei_2013_Modelling}. \ac{ILC} {is well-known for its ability to improve the tracking performance when the task repeats itself}  \cite{Arimoto_Kawamura_Miyazaki_84_BetteringOperationByLearning,Xu_Tan_03_NonlinearILC,Bristow_Tharayil_Alleyne_06_Survey_ILC,Chi_2015_Unified}. {Traditionally, ILC updates control actions at each sampling instant within the given time interval with appropriate convergence conditions to guarantee convergence}\cite{Arimoto_Kawamura_Miyazaki_84_BetteringOperationByLearning,Xu_Tan_03_NonlinearILC}. 

	Hence, in this work, \ac{RC} is used to formulate the reference tracking in the operational space with the consideration of constraints in the tri-space. The resultant problem is then transformed into an optimization problem with an unknown parameter set to balance the control effort and avoidance of the constraints. Additionally, the projection of the joint space acceleration  onto its null space is parameterized by another set of unknown parameters. The role of \ac{ILC} is thus to learn two sets of unknown parameters to improve the performance when the task is repetitive. By carefully choose initial values of parameters in these two sets, the feasible solutions of control of CDPRs are guaranteed. In order to reduce the ILC computational cost, this work utilizes novel two set parameterizations. One set balances the control effort and the satisfaction of constraints. The other set is the best parameterization in its null space. Moreover, the well-known optimization techniques such as \textit{\ac{PS}} and \textit{\ac{PSO}} are used to find optimal parameters iteratively. 
	
	{The main contributions of this work are summarized as follows:}
	\begin{enumerate}
		\item {A novel CDPR tri-space operational control framework is proposed. {This framework can track the reference trajectory in task space, handle kinematic and actuation redundancy, avoid cable-link interference, prevent loss of manipulability, and avoid reaching mechanical joint limits simultaneously.} To the best of the authors' knowledge, it is the first time that such a control framework is developed to handle multi-level redundancies and various constraints for CDPR.}
		\item The proposed framework combines RC, which provides feasible and optimal performance in the operational space and constraints in tri-space,  with ILC to find two sets of optimal parameters. One set tries to balance the control effort and satisfy constraints while the other set is the parameterization in the null space. {The proposed framework is thus able to improve the task performance over iterations with a low computational cost.}
		\item The capabilities of the control framework are illustrated through simulations of various operational space trajectories using the \textit{\ac{CASPR}} software \footnote{All methods and algorithms in this work has been implemented in the open-source cable robot software CASPR and can be accessed from: \texttt{http://www.github.com/darwinlau/CASPR}}  \citep{Lau_Eden_Tan_Oetomo_16_CASPR} on a \ac{BMArm} robot (2-link \ac{MCDR}), and the SpiderArm and FASTKIT-Planar HCDRs. Experiments on the BMArm are also conducted to show the robustness of the proposed framework.
	\end{enumerate}
	
	The remainder of the paper is organized as follows: Section  \ref{sec:RelatedWork} discusses the related work. Section \ref{sec:CDPR_Modeling} provides important notations, generalized MCDR and HCDR models with examples, and control objective. Section \ref{sec:OSFramework} discusses multiple CDPR control challenges and outlines the tri-space operational control framework. Sections \ref{sec:Reactive} and \ref{sec:ilc} proposes the  reactive and \ac{ILC} components, respectively. Section \ref{section:stability} presents the control framework stability analysis. Simulation and hardware results are presented in Sections \ref{sec:sim_results} and \ref{sec:hardware_results}, respectively. Finally, Section \ref{sec:conclusion} concludes the paper and discusses future work.
	
	\section{{Related Work}}\label{sec:RelatedWork}
	
	This section revisits the literature about handling redundancies of CDPRs. A lot of work has focused on \ac{SCDR}, which only involved a redundancy between joint and operational space. For example, in \cite{Lamaury_Gouttefarde_Chemori_Herve_13_DualAdaptiveControl}, a dual-space control scheme was proposed. The uncertainties in the dynamic models make an accurate feedforward control impossible. Consequently, the proposed tension
	distributions algorithm cannot guarantee a feasible solution
	within the entire robot workspace. To deal with disturbances appearing in the un-actuated joints, \cite{Qi_Khajepour_21_Redundancy} proposed a minimum actuated joint torque control for the actuated joints while resolving the kinematic redundancy without considering the tracking performance in joint space and the constraints. 
	
	In \cite{Zhang_2017_Tricriteria}, {a tricriteria optimization-coordination-motion scheme was proposed for dual-redundant-robots, which combined minimum velocity norm, repetitive motion planning, and infinity-norm velocity minimization solutions into a single QP formulation. In order to solve such a \textit{\ac{QP}} problem, different techniques were proposed. For example, the parameter-varying based neural networks were proposed in}\cite{Zhang_2018_Varying_Differential, Zhang_2018_Varying_Recurrent}. In \cite{Zhang_2018_Compatible}, {the training-free NN-based technique was proposed to deal with both convex and nonconvex sets while removing the initial error and accumulated error issues for redundant robots.} In \cite{Zhang_2020_Mutual}, {a mutual collision avoidance scheme was proposed by using line segment based distance measurement algorithm to solve the motion planning problem of dual manipulators. The above techniques have been shown to be efficient and accurate for solving dual-redundancy problem. However, how to balance the tracking accuracy, control effort and satisfaction of constraints has not been discussed. This is important otherwise tracking becomes unstable for some joint configuration, and eventually diverges} (see Section \ref{sec:react_bmarm}).
	
	Another common approach is the \emph{cascading} method (or two-stage optimization). The first stage determines a joint space trajectory that can track the reference by solving a serial robot \textit{\ac{IK}} problem. The second stage is to compute the cable forces by solving the CDPR \textit{\ac{ID}} problem \citep{Fang_Franitza_Torlo_Bekes_Hiller_04_Control_Optimal_Distribution, Lau_Oetomo_Halgamuge_15_JointInteractionInverseDynamics}.
	As the two stages are independent of each other, there might be some conflicts between them, i.e., the reference trajectory may not be feasible in the actuation space as pointed out in \citep{Khatib_87_UnifiedOpSpaceFormulation}.  In\cite{Qi_2019_Decoupled}, decoupled horizontal and vertical vibration models were developed, leading to a model predictive control of an HCDR. Such a technique can reduce trajectory tracking errors and vibrations without exploring the kinematic redundancy. In \cite{Michelin_21_Path}, standard kinematic resolution techniques based on local optimization of the null space component were used to generate feasible generalized coordinates for HCDR end-effector tracking task. However, the work only focused on the kinematic redundancy, and not on the actuation redundancy. 
	
	In \cite{DeSapio_Warren_Khatib_Delp_05_SimulatingTaskLevelControl}, Khatib's operational space control \citep{Khatib_87_UnifiedOpSpaceFormulation} was extended to determine the muscle forces required to achieve reactive operational space control of biomechanical systems in which the workspace is conservatively approximated by a set of linear constraints \citep{DeGroot_Brand_01_RegressionShoulder}. As such, the control command can be determined by solving convex QP. However, for CDPRs, the linear constraints approximation of the pre-generated workspace is not practical due to the high workspace irregularity and computational costs for high DoF robots. As such, the approach proposed in \cite{DeSapio_Warren_Khatib_Delp_05_SimulatingTaskLevelControl} for biomechanical systems cannot be effectively used for CDPRs in general. {Hence, in summary, there is no effective solution for a tri-space operational control to deal with two-level  redundancy problem with the consideration of  feasibility in the CDPRs, which motivates the proposed tri-space control framework for CDPRs.}
	\section{Notations, CDPR Modelling and Control Objective}\label{sec:CDPR_Modeling}
	
	This section provides relevant mathematical notations, the generalized modeling of CDPRs including MCDRs and HCDRs, followed by the control objective of this work.
	
	\subsection{Notations}
	Let $\mathbb{N}$ represent the set of all non-negative integers. The notation $\mathbb{R}$ represents the set of all real numbers, $\mathbb{R}_{\geq 0}$ is the set of all non-negative real numbers, and $\mathbb{R}_{>0}$ is the set of all positive real numbers. 
	For any vector $\bm{x} \in \mathbb{R}^n$, $\norm{\bm{x}}_2$ represents its Euclidean norm,
	defined as $\norm{\bm{x}}_2\triangleq \sqrt{ \bm{x}^T\bm{x}}$, where $(\cdot)^T$ represents the transpose.  For two vectors in $\mathbb{R}^{n}$ with $\bm{x}=\left[\begin{array}{ccc}x_1&\cdots&x_n\end{array}\right]^T$ and $\bm{y}=\left[\begin{array}{ccc}y_1&\cdots&y_n\end{array}\right]^T$, $\bm{x}\leq \bm{y}$ indicates that $x_i\leq y_i,$ $\forall$ $i=1,\ldots, n$.
	
	For a matrix $A=\left\{a_{i,j}\right\}\in \mathbb{R}^{n\times m}$, $\norm{A}_2$ is the induced matrix norm defined as $\norm{A}_2=\sqrt{A^TA}$. The Frobenius norm of $A$ is defined as $\norm{A}_F=\sqrt{\sum_{i=1}^n\sum_{j=1}^m a_{i,j}^2}$. The pseudo-inverse $A^\dagger$ is defined such that $A A^\dagger = I_{n}$, where $I_n$ denotes an n-dimensional square identity matrix. The null space of matrix $A$ is the set $\mathcal{N}_A = \left\lbrace\mathbf{x} \in \mathbb{R}^m: A\mathbf{x} = \mathbf{0} \right\rbrace$. The matrix  $N_A \in \mathbb{R}^{m \times (m - n)}$ denotes a matrix whose columns form a basis of the set $\mathcal{N}_A$. 
	
	\subsection{SCDR and MCDR Modelling}
	
	A \ac{MCDR} is a type of \ac{CDPR} where the number of links is more than one. If the number of links is one then the MCDR becomes a \ac{SCDR}. Let the joint space $\bm{q}=[q_1,\cdots,q_{n_M}]^{T}\in \mathbb{R}^{n_M}$ is defined as the generalized coordinates. The cable lengths and forces are denoted as $\bm{l}=[l_1,\cdots,l_{m_M}]^{T} \in \mathbb{R}^{m_M}$ and $\bm{f}=[f_1,\cdots,f_{m_M}]^{T} \in \mathbb{R}^{m_M}$, respectively.
	Hence, the kinematic relationship between $\bm{l}$ and $\bm{q}$ is described by
	\begin{equation}
		\bm{l} = \bm{s}_1(\bm{q}),
		\label{eqn:IK}
	\end{equation}
	where the nonlinear mapping $\bm{s}_1: \mathbb{R}^{n_M} \rightarrow \mathbb{R}^{m_M}$ is the inverse kinematics (IK) function \citep{Lau_Oetomo_Halgamuge_13_GeneralisedMultilinkCRM}. The derivative of \eqref{eqn:IK} gives
	\begin{equation}
		\dot{\bm{l}} = \bm{L}(\bm{q})\dot{\bm{q}},
		\label{eqn:IK_J}
	\end{equation}
	where $\bm{L}(\bm{q}) =[{\partial \bm{s}_1(\bm{q})}/{\partial q_1},\cdots,{\partial \bm{s}_1(\bm{q})}/{\partial q_{n_M}}]\in \mathbb{R}^{m_{M} \times n_{M}}$ is the \textit{joint-cable Jacobian matrix}. Due to the unilateral actuation property of cables, \emph{actuation redundancy} ($m_M \geq n_M + 1$) is needed  to produce motion in all DoF. The dynamics of the system is described by the following \textit{\ac{EoM}}
	\begin{IEEEeqnarray}{C}
		\bm{M}(\bm{q})\ddot{\bm{q}}+\bm{C}(\dot{\bm{q}},\bm{q}) + \bm{G}(\bm{q})= -\bm{L}(\bm{q})^T\bm{f},\label{eqn:eom_mcdr}\\
		\bm{0} \leq \bm{\underline{f}} \leq \bm{f} \leq \bm{\overline{f}}.
		\label{eqn:f_bounds}
	\end{IEEEeqnarray}
	The left hand side of \eqref{eqn:eom_mcdr} refers to the \textit{system wrench}, composed of the \textit{mass-inertia matrix} $\bm{M}(\bm{q})\in \mathbb{R}^{n_M\times n_M}$, the \textit{centrifugal and Coriolis vector} $\bm{C}(\dot{\bm{q}},\bm{q})\in \mathbb{R}^{n_M}$, and the \textit{gravitational vector} $\bm{G}(\bm{q}) \in \mathbb{R}^{n_M}$. The \textit{joint-cable Jacobian matrix} $\bm{L}(\bm{q})$ maps the cable force vector $\bm{f}$ into the joint space system wrench. Cable forces are requested to be bounded by the minimum and maximum positive force bounds $\bm{\underline{f}}$ and $\bm{\overline{f}}$ in (\ref{eqn:f_bounds}), respectively.
	
	The MCDR operational space can be defined as the position and/or orientation of the end-effector $\bm{x}=[x_1,\cdots,x_r]^{T} \in \mathbb{R}^r$, where $r$ refers to the number of operational space \ac{DoF}. Usually $r\ll n_M$. For this work, Euler angles are chosen over quarternions because quarternions are much less intuitive than Euler angles. However, since the framework is generic, hence quaternions can also be used to represent the operational space. The kinematic relationship between the joint and operational space (Euler space) can be written as
	\begin{equation}
		{\bm{x}} = \bm{s_2}({\bm{q})}.
		\label{eqn:os_0}
	\end{equation}
	In \eqref{eqn:os_0}, $\bm{s}_2:\mathbb{R}^{n_M}\rightarrow\mathbb{R}^{r}$ represents the \ac{FK}, mapping the joint space to operational space $\mathbb{R}^{r}$. The derivative of \eqref{eqn:os_0} becomes
	\begin{equation}
		\dot{\bm{x}} = \bm{J}(\bm{q})\dot{\bm{q}},
		\label{eqn:os}
	\end{equation}
	where $\bm{J}(\bm{q}) =[{\partial \bm{s}_2}(\bm{q})/{\partial q_1},\cdots,{\partial \bm{s}_2}(\bm{q})/{\partial q_{n_M}}] \in \mathbb{R}^{r \times n_M}$ is the \emph{joint-operational Jacobian} matrix. This leads to the following derivative of (\ref{eqn:os}):
	\begin{equation}
		\ddot{\bm{x}} = \bm{J}(\bm{q})\ddot{\bm{q}} + \dot{\bm{J}}(\bm{q})\dot{\bm{q}}.
		\label{eqn:os_J}
	\end{equation}
	Eq \eqref{eqn:os_J} can be re-written as
	\begin{equation}
		\ddot{\bm{q}} = \bm{J}^{\dagger}(\bm{q})(\ddot{\bm{x}}-\dot{\bm{J}}(\bm{q})\dot{\bm{q}})+\ddot{\bm{q}}_{N},
		\label{eqn:os_J_2}
	\end{equation}
	where $ \ddot{\bm{q}}_{N} $ and $ {\bm{J}}^{\dagger} $ are null space vector and pseudoinverse of $\bm{J}$, respectively. 
	
	\subsection{HCDR Modelling}
	\acp{HCDR} are hybrid actuated systems that combine cables and other types of rigid-link actuators. For an HCDR with $m$-cables with its CDPR joint space $\bm q_M \in \mathbb{R}^{n_M}$  and $p$-actuators, the original CDPR joint space $\bm q_M$, is extended to include the rigid-link actuated \ac{DoF} $\bm{q}_a \in \mathbb{R}^p$. As a result, the joint space of the HCDR becomes $\bm{q} = [{\bm{q}}_M^T ~~ \bm{q}_a^T ]^T \in \mathbb{R}^n$, where $n= n_M + p$. The kinematic relationship between the cable lengths $ \bm{l} $, and  CDPR joint space $ {\bm{q}}_M $ is the same as that of \eqref{eqn:IK}, and \eqref{eqn:IK_J} with appropriate dimensions. The actuation commands $\bm{a}$ is formed by combining the cable force $\bm{f} \in \mathbb{R}^{m}$ and the active joint torque $\bm{\tau} \in \mathbb{R}^{p}$ as:
	\begin{equation}
		\bm{a} =
		\left[\begin{array}{c}
			\bm{f} \\ \bm{\tau}
		\end{array}\right] := \left[\begin{array}{c}
			\bm{a}_C \\ \bm{a}_D
		\end{array}\right]\in \mathbb{R}^{m+p}. \label{eqn:a}
	\end{equation}
	The bounds on $\bm{a}$ are defined as
	\bea
	\bm{\underline{a}} \leq \bm{a} \leq \bm{\overline{a}}, \nonumber \\
	\bm{\underline{a}} =
	\left[\begin{array}{c}
		\bm{\underline{f}} \\ \bm{\underline{\tau}}
	\end{array}\right], \quad
	\bm{\overline{a}} =
	\left[\begin{array}{c}
		\bm{\overline{f}} \\ \bm{\overline{\tau}}
	\end{array}\right], \label{eqn:a_bound_2}
	\eea
	where $\bm{\underline{\tau}}, \bm{\overline{\tau}} \in \mathbb{R}^{p}$ are the minimum and maximum joint torque. 
	
	The dynamics of HCDRs can be expressed by the following extended \ac{EoM} from \eqref{eqn:eom_mcdr} and \eqref{eqn:f_bounds}
	\bea
	\bm{M}(\bm{q})\ddot{\bm{q}}+\bm{C}(\dot{\bm{q}},\bm{q}) + \bm{G}(\bm{q})= \bm{B}(\bm{q})\bm{a},  \label{eqn:eom_general}\\
	\bm{B}(\bm{q})=
	\left[\begin{array}{cc}
		-\bm{L}(\bm{q})^{\bm{T}}& \bm{A}(\bm{q})
	\end{array}\right], \label{B_def}\\
	\bm{\underline{a}} \leq \bm{a} \leq \bm{\overline{a}},\label{constraints}
	\eea
	where $\bm{B}(\bm{q})\in \mathbb{R}^{n\times (m+p)}$ is formed by the {joint-actuator} Jacobian matrix $ \bm{L}(\bm{q}) $, and the Jacobian matrix for active joint torques $\bm{A}(\bm{q})\in \mathbb{R}^{n\times p}$. When $p=0$, the extended EoM of HCDRs defined by (\ref{eqn:eom_general}) reduces to the EoM for MCDRs or SCDRs in (\ref{eqn:eom_mcdr}). Therefore, the extended EoM given by (\ref{eqn:eom_general}) will be used to describe the dynamics of CDPRs in the remainder of the paper.
	
	\subsection{Control Objective}
	
	It is noted that most tracking tasks are defined in the operational space, but due to redundancy coming from CDPRs (referring to MCDRs and HCDRs), a good operational tracking performance does not guarantee feasible joint space performance. When the actuation space is considered to minimize the control effort, the control design becomes more challenging.  Additionally, undesirable situations such as cable-link interference, loss of manipulability, and reaching mechanical joint limits might happen. 
	The aim of this work is thus to exploit the two-level redundancy of CDPRs to achieve good performance in the operational and joint space while minimizing
	the control effort and avoiding undesirable situations simultaneously. More precisely, the control objective is summarized as:
	
	For a given desired trajectory $\bm{x}_d(t)\in  \mathbb{R}^{r}$, $ t\in [0,T]$ for some $T\in \mathbb{R}_{>0}$, the control objective of a CDPR is to track this desired operational space trajectory with the desired joint space performance with simultaneous minimized control effort and avoidance of undesirable situations, when the  task performs repetitively.
	
	\section{Proposed Reactive-Iterative Tri-Space Operational Control Framework}\label{sec:OSFramework}
	
	This section presents a robust control framework to achieve the control objective for CDPRs. For given $\bm{x}_d(t)$, from (\ref{eqn:os_J_2}), the joint space motion $\bm{q}_d(t)$ satisfies:
	\bea
	\ddot{\bm{q}}_d = \bm{J}^{\dagger}(\bm{q}_d)(\ddot{\bm{x}}_d-\dot{\bm{J}}(\bm{q}_d)\dot{\bm{q}}_d)+\ddot{\bm{q}}_{N}.
	\label{joint_space_desired}
	\eea
	Equation \eqref{joint_space_desired} composed of two parts: one is the desirable joint space trajectory $\bm{q}_d(t)$ and the other is related to the null space trajectory ${\bm{q}}_{N}(t)$. Due to the kinematic redundancy, the solution of (\ref{joint_space_desired}) is not unique. Optimization techniques are widely used to solve such a redundancy problem,.
	
	The cost function used in the optimization is based on the widely used  lower-level \textit{Proportional-Derivative (PD)} control law, performance requirements in CDPR (MCDR and HCDR), as well as a special form of parameterization of the null space trajectory ${\bm{q}}_{N}$. 
	
	\subsection{PD Control Law}
	
	For a given reference trajectory in operation space $\bm{x}_d(t)\in\mathbb{R}^{r}$,  a PD controller has been widely used. Let the tracking error be 
	\bea
	\bm{e}(t)=\bm{x}_d(t) - \bm{x}(t),\label{error}
	\eea
	where $\bm{x}$ is the actual operational space pose as defined in (\ref{eqn:os_0}).  An ideal PD controller can eliminate the nonlinearities of the CDPR, leading to the following ``ideal'' closed-loop system:
	\begin{align}
		(\ddot{\bm{x}}_{d}-\ddot{\bm{x}} )  + \bm{K}_d (\dot{\bm{x}}_{d}-\dot{\bm{x}} ) + \bm{K}_p (\bm{x}_{d}-\bm{x}) = \bm{0}_{r}, \label{eqn:e_t}
	\end{align}
	or equivalently, 
	\begin{align}
		\ddot{\bm{e}}+ \bm{K}_d \dot{\bm{e}} + \bm{K}_p {\bm{e}} = \bm{0}_{r}, \label{eqn:e_t_2}
	\end{align}
	where $\bm{K}_d, \bm{K}_p \in \mathbb{R}^{r\times r}$ are diagonal matrices with positive diagonal elements.  For the convenience of notation, we defined the set $\mathcal{X}_d(t) = \{\bm{x}_{d}(t), \dot{\bm{x}}_{d}(t), \ddot{\bm{x}}_{d}(t)\}$ containing the desired trajectory and its derivatives. Moreover, it is denoted 
	\begin{align}
		\bm{b}(\mathcal{X}_d,\bm{q},\dot{\bm{q}},\bm{e}) = \ddot{\bm{x}}_{d} - \dot{\bm{J}}\dot{\bm{q}} + \bm{K}_d \dot{\bm{e}}+ \bm{K}_p {\bm{e}}. \label{eqn:os_error_b}
	\end{align}
	\begin{remark}
		{It is noted that the choice of PD control matrices $\bm{K}_d$ and $\bm{K}_p$ will not affect the stability properties of the ``ideal'' closed-loop, though their choices affect the transient behaviours. Even though the CDPR is working in non-ideal situations, the existence of PD control law will ensure the boundedness of the operation trajectories $\bm{x}(t)$ for any $t\in [0,T]$ as shown in Theorem 2. The introduction of \ac{ILC} for updating the parameters will improve the performance over iterations when the CDPR  is performing a task repetitively. Hence, the choice of PD control matrices is not the focus of this work. In our analysis, these matrices are fixed. \hfill $\circ$}
	\end{remark}
	Substituting $\ddot{\bm{x}} =\ddot{\bm{x}}_{d} + \bm{K}_d \dot{\bm{e}}+ \bm{K}_p {\bm{e}}$ into \eqref{eqn:os_J_2}, and writing $\ddot{\bm{q}}$ as joint acceleration command $\ddot{{\bm{q}}}_{c}(\mathcal{X}_d, \bm{q}, \dot{\bm{q}},\bm{e})$, \eqref{eqn:os_J_2}  can be re-written as
	\begin{equation}
		\ddot{{\bm{q}}}_{c}(\mathcal{X}_d,\bm{q},\dot{\bm{q}},\bm{e}) = \bm{J}^{\dagger}\bm{b}(\mathcal{X}_d,\bm{q},\dot{\bm{q}},\bm{e})+\ddot{\bm{q}}_{c_N}.
		\label{eqn:os_J_3}
	\end{equation}
	It is noted that if $\ddot{\bm{q}}$ is computed, both $\dot{\bm{q}}$ and ${\bm{q}}$ can be obtained using numerical integration techniques. Furthermore, $\bm{e}$ is defined in operation space instead of joint space. Thus we can re-write $\ddot{ {\bm{q}}}_{c}(\mathcal{X}_d,\bm{q},\dot{\bm{q}},\bm{e})$ as $\ddot{ {\bm{q}}}_{c}(\mathcal{X}_d,\ddot{\bm{q}})$ and $\bm{b}(\mathcal{X}_d,\bm{q},\dot{\bm{q}},\bm{e})$ as $\bm{b}(\mathcal{X}_d,\ddot{\bm{q}})$ when optimization is designed in joint space. In an ideal case, if the operational space errors are fully compensated (i.e. $ \bm{e} =  \bm{\dot e} = 0$), \eqref{eqn:os_J_3} takes the form of \eqref{joint_space_desired}.
	
	Let  $\bm{J}^\dagger_W = \bm{W}_q^{-1}\bm{J}^T(\bm{J}\bm{W}_q^{-1}\bm{J}^T)^{-1} $ be the weighted pseudoinverse of $\bm{J}$ \citep{Whitney_69_resolvedmotionrate} with a symmetric positive definite weighting matrix $\bm{W}_q \in \mathbb{R}^{n \times n}$. The use of $\bm{W}_q$ is suitable when the joint space of the CDPR has mixed units, such as between translation and rotations. Replacing $\bm{J}^{\dagger}$ by $\bm{J}_{W}^{\dagger}$ in \eqref{eqn:os_J_3}, yields:
	\bea
	\ddot{{\bm{q}}}_{c}(\mathcal{X}_d, \ddot{\bm{q}}) = \bm{J}_{W}^{\dagger}\bm{b}(\mathcal{X}_d,\ddot{\bm{q}})+\ddot{\bm{q}}_{c_N}.
	\label{EOM_final}
	\eea
	When the joint space trajectory $\bm{q}(t)$ is designed, it is expected that $\bm{q}(t)$ can track the commanding joint acceleration $\ddot{ {\bm{q}}}_{c}(\mathcal{X}_d)$ with the consideration of other joint space requirements. The error between $ \bm{\ddot q}(t) $ and $\ddot{ {\bm{q}}}_{c}(\mathcal{X}_d)$ can be formulated as one optimization problem.
	
	\subsection{Cost Function for Joint Space Trajectories}
	
	The user-defined cost function is denoted as $V(\mathcal{X}_d, \ddot{\bm{q}}, \bm{a})$ is related to control objectives. In this work, it comprises of three quadratic components: 1) \emph{operational space tracking function} $g_t(\mathcal{X}_d, \ddot{\bm{q}})$ resolving the kinematic redundancy; 2) \emph{actuation function} $g_f(\bm{a})$ minimizing control effort and resolving the actuation redundancy and; 3) \emph{the function related to constraints or avoidance function} $g_a(\ddot{\bm{q}})$, to which aims undesirable situations. Hence, $V(\mathcal{X}_d, \ddot{\bm{q}}, \bm{a})$ can be expressed as
	\bea
	V(\mathcal{X}_d, \ddot{\bm{q}}, \bm{a})= g_t(\mathcal{X}_d, \ddot{\bm{q}})
	+ \alpha \cdot g_f(\bm{a}) + \beta \cdot g_a(\ddot{\bm{q}}),
	\label{eqn:qp_obj}
	\eea
	where $\alpha> 0$ and $\beta > 0$ are scaling factors for the actuation function and avoidance function, respectively. As a general guideline, $\alpha \ll \beta <1$ is selected to prioritize tracking over actuation and avoidance function. Other forms of cost function can be used. The positive pair $(\alpha,\beta)$ is learnt when the CDPR performs repetitive tasks.
	
	The details of choosing the cost function $V(\mathcal{X}_d, \ddot{\bm{q}}, \bm{a})$ and other constraints in joint space will be discussed in Section \ref{sec:Reactive}. The \ac{RC} will be used to determine an appropriate joint space trajectory $\ddot{\bm{q}}(t)$ for $t\in [0,T]$.

	\begin{figure*}[htb!]
		\centering
		\def\svgwidth{\fs\textwidth}
		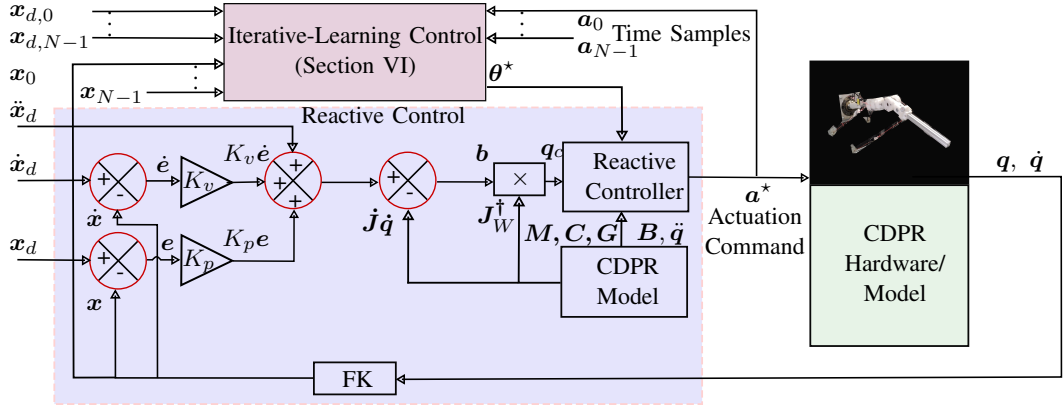
		\caption{Tri-space operational control framework.}
		\label{fig:osframework0}
	\end{figure*}
	
	\subsection{Parameterization and its Cost}
	
	Since the null space vector $\ddot{\bm{q}}_{c_N}$ plays an important role in joint space optimization, in this work, a novel parameterization of the null space vector $\ddot{\bm{q}}_{c_N}$ is considered such that \eqref{EOM_final}
	can be re-written as 
	\bea
	\bm{\ddot{{ q}}}_{c_N}= (\bm{I} - \bm{J}_W^\dagger\bm{J}) \bm{D}\bm{J}_{W}^{\dagger}\bm{b}(\mathcal{X}_d,\ddot{\bm{q}}),
	\label{eqn:q_c_N}
	\eea
	where $\bm{D}=diag(d_1,\cdots,d_n) \in \mathbb{R}^{n \times n}$ is an unknown diagonal matrix that scales the components of $\bm{J}_{W}^{\dagger}\bm{b}(\mathcal{X}_d,\ddot{\bm{q}})$, which is an $ n $-dimensional vector. The columns of the matrix $\bm{D}$ need to be identified for improving the performance of the CDPR over repetitive tasks. The matrix $\bm{I} - \bm{J}_W^\dagger\bm{J}$ is the null space projection matrix. Consequently, \eqref{EOM_final} is re-written as
	\bea
	\ddot{{\bm{q}}}_{c}(\mathcal{X}_d, \ddot{\bm{q}}) = \bm{J}_{W}^{\dagger}\bm{b}(\mathcal{X}_d,\ddot{\bm{q}})+(\bm{I} - \bm{J}_W^\dagger\bm{J}) \bm{D}\bm{J}_{W}^{\dagger}\bm{b}(\mathcal{X}_d,\ddot{\bm{q}}).
	\label{EOM_final_parameterization}
	\eea
	
	Together with $\alpha$ and $\beta$ from \eqref{eqn:qp_obj}, a reactive tuning parameter vector $\btheta \in \mathbb{R}^{n+2}$ can be defined as
	\bea
	\btheta=\left[d_1, \cdots, d_n, \ln(\alpha), \ln(\beta)\right]^T. \label{theta}
	\eea
	The parameter $\btheta$ needs to be identified through repetition. Hence, the widely used technique of \ac{ILC}, which improves the task performance through iterations when the trajectory is repeatedly executed, is used to learn an optimal $\btheta^*$.
	\begin{remark}
		{{It is highlighted that in this framework, two set of optimal parameters are sought. The first set tries to balance the control effort and satisfaction of constraints, while the second optimal parameters are related to parameterization in the null space, i.e., the best parameters in the null space to achieve the optimal performance in operational space. More discussions on null space parameterization are presented in} Section \ref{sec:ilc}.
		\hfill $\circ$}
	\end{remark}

	\subsection{Overall Structure}\label{section:overall}
	
	The proposed tri-space operational control framework consists of two primary components:
	\begin{enumerate}
		\item \acf{RC} plans the joint space trajectory to solve redundancy of CDPR so that it can track the  desired trajectory $\bm{x}_d(t), t\in [0,T]$ in operational space and achieve desirable performance in terms of $\ddot {\bm{q}}(t)$ with respect to the cost $V(\mathcal{X}_d, \ddot{\bm{q}}, \bm{a})$ and other constraints in joint space. Such a problem can be converted into an optimization problem with appropriate constraints if $\ddot {\bm{q}}(t)$ and $\bm{a}(t)$ are sampled with the sampling rate $T_s$.  More precisely, at each sampling instant $t_k=0,1,\ldots, N-1$, an optimization problem is solved. Here  $t_k=kT_s, k=0,1,\ldots,N-1$, with $T_s$ satisfying $N=\frac{T}{T_s}$. A standard \ac{QP} controller is used to solve both the kinematic and actuation redundancies (tri-space redundancies) simultaneously at each sampling instant. The RC is responsible for maintaining the system stability and satisfying the unilateral constraints of $\bm{a}(t)$, when encountering new trajectories without any past experiences. The reactive controller also operates real-time at each time instant (\emph{faster-time scale}). More details of the RC in Section \ref{sec:Reactive} as well as Algorithm 2.

		\item \acf{ILC} improves the task performance through iterations by selecting the optimal parameter $\btheta^\star$ with respect to some user-defined cost function $P(\cdot)$, where $\btheta$ is defined in \eqref{theta} when the trajectory is repeatedly executed (Section \ref{sec:ilc}). To avoid confusion, the ILC cost function is referred to as the trajectory performance function. The ILC focuses on improving the performance through exploring the joint space redundancy when the operational space task is repeated. The ILC performs each update at the end of each trajectory cycle (\emph{iteration}) at a \emph{slower-time scale}.
		
	\end{enumerate}
	The overall structure of Tri-space operational control framework is provided in Fig. \ref{fig:osframework0}, where the implementation details of the \ac{RC} are shown comprehensively. The ILC implementation details are discussed in Section \ref{sec:ilc}. The operation modes of the control framework are as follows:
	
	\subsubsection{Hardware Mode}
	In the controller operation, both \textit{hardware} and \textit{simulation} modes are available. In hardware mode, the length commands are sent to the BMArm through \ac{ROS} at $200$ Hz and cable lengths are controlled by the myomuscle units \citep{Marques_Maufroy_Lenz_Dalamagkidis_Culha_Siee_Bremner_13_MYORobotics}. The length commands are generated from the optimized actuation command $ \bm{a}^{\star}(t_k) $, by applying a Forward Dynamic (FD) model. At each sampling instant $ t_k $, joint space velocity $\dot{\bm{q}}{(t_k)}$, and position $\bm{q}({t_k})$ are determined through FK, using the cable length feedback, calculated from the motor encoder feedback (Fig. \ref{fig:osframework0}). This FK model is located within the CDPR/hardware block. As shown in Fig. \ref{fig:osframework0}, the operational space position $\bm{x}({t_k})$ and velocity $\dot{\bm{x}}({t_k})$ are evaluated through FK with $\bm{q}({t_k})$ and $\dot{\bm{q}}({t_k})$ computed using \eqref{eqn:os}. To validate the cable force commands, load cells were placed in-line with the cables to measure the cable forces $\bm{f}(t_k)$. 
	
	\subsubsection{Simulation Mode} For simulation mode, instead of the hardware, the length commands are sent to a CASPR CDPR model. Control frequencies of 100 and 200 Hz were selected for various simulations. During simulations, to simulate the sensor noise, additive noise was added to the model's length feedback to generate the cable lengths $ {\bm{l}(t_k)} $.

	\section{Reactive Controller}\label{sec:Reactive}
	At each sampling instant $t_k$, the CDPR has joint information $\bm{q}_k:={\bm q}(t_k)$, $\dot {\bm{q}}_k:=\dot{\bm q}(t_k)$, and $\ddot {\bm{q}}_k:=\ddot{\bm q}(t_k)$. This section will first formulate the \ac{RC} design as an optimization problem, followed by the discussion of the cost function $V(\mathcal{X}_d, \ddot{\bm{q}}, \bm{a})$ defined in (\ref{eqn:qp_obj}). Finally, Section \ref{sec:avoid} discusses the details of avoiding undesirable situations. 
	\subsection{QP Optimization Formulation}
	The primary objective of the reactive controller is to determine a set of feasible actuation commands $\bm{a}_k = {\bm a}(t_k) = [\bm{f}_k^{T}, \bm{\tau}_k^{T}]^T$ and the reference joint trajectory $\ddot{\bm{q}}_k$ that actuates the system to track the operational space trajectory $\mathcal{X}_d(t_k)$, while avoiding undesirable positions for $k=1,2,\ldots, N-1$ as defined by the cost function $V(\mathcal{X}_d, \ddot{\bm{q}}, \bm{a})$ with the consideration of cable force, joint torque and \ac{EoM} constraints coming from \eqref{eqn:a_bound_2} to \eqref{constraints}. This leads to the following convex \ac{QP} formulation, which deals with both levels of redundancy (kinematic and actuation) in the tri-space:
	
	\begin{mini}|l|
		{\ddot{\bm{q}_k}, \bm{a}_k}{ V\left(\mathcal{X}_d(t_k), \ddot{\bm{q}}_k, \bm{a}_k\right)}{\label{eqn:reactive_qp} }{}\hspace{0.5in}
		\addConstraint\\
		{\bm{M}(\bm{q}_k)\ddot{\bm{q}}_k+\bm{C}(\dot{\bm{q}}_k,\bm{q}_k) + \bm{G}(\bm{q}_k)= \bm{B}(\bm{q}_k)\bm{a}_k}
		\addConstraint{\bm{\underline{a}} \leq \bm{a}_k \leq \bm{\overline{a}}}
		\addConstraint{\bm{\Phi}\left [
			\begin{array}{c}
				\ddot{\bm{q}_k} \\ \bm{a}_k
			\end{array}
			\right ]
			\leq \bm{\rho},} 
	\end{mini}
	where $\bm{\Phi}\in \mathbb{R}^{n_q\times (m+n+p)}$ and $\bm{\rho}\in \mathbb{R}^{n_q}$ define the controller inequality constraints for some interger $n_q$. 
	\begin{remark}
		{The choices  of $\bm{\Phi}$ and $\bm{\rho}$ are  application dependent. In this work, the RC controller contains three types of linear constraints: the EoM \eqref{eqn:eom_general}, actuation bounds and other task constraints. The EoM ensures that the system dynamics can be satisfied, while the actuation bounds ensure the resulting actuation commands are feasible. Other inequality constraints are hard constraints that maintain the system's capability of achieving the tracking task (see Section \ref{sec:avoid}).} \hfill $\circ$
	\end{remark}
	
	Next we will discuss the operational space tracking $g_t(\cdot,\cdot)$ and actuation command $g_f(\cdot)$ terms of \eqref{eqn:qp_obj}. The $g_a(\cdot)$ term is discussed in Section \ref{sec:react_avoid}.

	\subsubsection{Operational Space Tracking}
	
	From \eqref{EOM_final}, a good candidate for $g_t(\mathcal{X}_d, \ddot{\bm{q}})$ is
	\bea
	g_t(\mathcal{X}_d, \ddot{\bm{q}}) 
	=\norm{\ddot{\bm{q}} - \bm{J}^\dagger_W\bm{b}(\mathcal{X}_d,\bm{\ddot q})}^2_2,  \label{eqn:g_t}
	\eea
	where $\bm{b}(\cdot,\cdot)$ is defined in (\ref{eqn:os_error_b}). Other choices of the tracking performance can be used. 
	
	\subsubsection{Actuation Command}
	
	The objective function component $g_f(\bm{a})$ is responsible for resolving the actuation redundancy for CDPR systems. As such, the actuation command $\bm{a}$ should: 1) produce the required joint acceleration $\ddot{\bm{q}}$ while satisfying the EoM \eqref{eqn:eom_general}; 2) be within the feasible actuation bound (\ref{eqn:a_bound_2}); and 3) satisfy some desired objectives such as minimal \emph{control effort}. The EoM and actuation bounds are represented as constraints in the QP (\ref{eqn:reactive_qp}), and the minimization of control effort can be simply expressed by the objective function component
	\begin{align}
		g_f(\bm{a}) = \hspace{1mm}\bm{a}^T \bm{W}_a\bm{a},
		\label{eqn:rqp_actuation_obj}
	\end{align}
	where $\bm{W}_a \in \mathbb{R}^{(m+p) \times (m+p)}$ is positive-definite, and typically a diagonal matrix for decoupling different joint actuation efforts in the controller. 
	
	The choice of $g_t(\cdot,\cdot)$ and $g_f(\cdot)$ was quite straightforward. In this section, from henceforth, the focus of the discussion is the avoidance of undesirable situations which forms the foundation for the $g_a(\cdot)$ term of \eqref{eqn:qp_obj} and also defines the controller inequality constraints $\bm{\Phi}$ and $\bm{\rho}$ of (25). 
	\subsection{Avoidance of Undesirable Situations}\label{sec:avoid}
	Despite the ability of the tracking task and resolution of the two-level redundancies, the system is not robust if the reactive controller does not consist of an avoidance function. To prevent the system from encountering undesirable situations, such as loss of manipulability, cable interference, and reaching mechanical joint limits.
	
	To avoid such problems, a combination of hard and soft constraints within the reactive controller \eqref{eqn:reactive_qp} is used to avoid undesirable situations. In addition to the loss of manipulability, other undesirable situations include the interference of cables and also limits on the range of joint motion. {In this work, the function $h(\bm{q})$ is used to represent either constraints or the cost such as}
	\begin{itemize}
		\item Satisfaction of $h(\bm{q}) \geq h_{min}$ as a hard constraint
		\item Increase of $h(\bm{q})$ as an objective to avoid $h_{min}$
	\end{itemize}
	
	\subsubsection{Hard linear constraints}
	
	In this work,  the undesirable situation is represented as  $h_{min} \leq h(\bm{q}) \leq h_{min} + \Delta_h$, where $\Delta_h \in \mathbb{R}^+$ represents the region size that is considered to be close to the undesirability. Since $h(\bm{q})$ is nonlinear, it is proposed to add a velocity level constraint when $h_{min} \leq h(\bm{q}) \leq h_{min} + \Delta_h$ 
	\begin{align}
		\dot{h}(\bm{q}) = \frac{\partial h(\bm{q})}{\partial \bm{q}} \dot{\bm{q}} \geq \epsilon, \label{eqn:react_hardcon_q_d}
	\end{align}
	where $\epsilon > 0$ is the minimum increase in $h(\bm{q})$ required such that the system would move away from the constraint. As the optimization variable of the QP involves $\ddot{\bm{q}}$ instead of $\dot{\bm{q}}$, the velocity can be expressed numerically as
	\begin{align}
		\dot{\bm{q}} = \dot{\bm{q}}_{p} + \ddot{\bm{q}} T_s, \label{eqn:q_diff}
	\end{align}
	where $T_s$ is the time step and $\dot{\bm{q}}_{p}$ is the joint space velocity at the previous time step. By combining (\ref{eqn:react_hardcon_q_d}) with \eqref{eqn:q_diff}, the following linear constraint can be included in RC to avoid undesirable situations
	\begin{align}
		\bm{\Phi} = \left[ -T_s \frac{\partial h}{\partial \bm{q}} \quad \bm{0}_{m+p} \hspace{1ex} \right] ,
		\quad
		\bm{\rho} = \frac{\partial h}{\partial \bm{q}}\dot{\bm{q}}_{p} - \epsilon~,
		\label{eqn:react_hardcon_q_dd}
	\end{align}
	where $\bm{0}_{m+p}$ is an $m+p$ zero row vector. Since ${\partial h(\bm{q}, \dot{\bm{q}})}/{\partial \bm{q}}$ is a constant row vector at a particular state $\bm{q}$ so (\ref{eqn:react_hardcon_q_dd}) is a linear constraint.
	
	\subsubsection{Avoidance function} \label{sec:react_avoid}
	
	Hard constraints are activated such that the system can react when it is too close to undesirable situations or even failure. However, the activation of hard constraints creates sudden changes of acceleration, which potentially creates non-smooth joint space motions. Hence, in addition to hard constraints, an avoidance function $g_a(\ddot{\bm{q}})$ is proposed to guide the system smoothly away from undesirable situations prior to reaching the hard constraints, increasing the capability to avoid undesirable situations in advance. The avoidance function $g_a(\ddot{\bm{q}})$ of \eqref{eqn:qp_obj} can be formulated as
	\begin{align}
		g_a(\ddot{\bm{q}}) = \norm{\ddot{\bm{q}} -  {\ddot{\bm{q}}}_A}^2_2,  \label{eqn:avoid_fun_general}
	\end{align}
	where ${\ddot{\bm{q}}}_A$ refers to the \emph{avoiding acceleration}, {which is the required acceleration to drive the system to avoid undesirable situations.} The avoidance objective can be formulated as a desired velocity in order to increase the gradient of function $h(\bm{q})$ and the desired acceleration ${\ddot{\bm{q}}}_A$ expressed as
	\begin{align}
		{\ddot{\bm{q}}}_A =  \left(k_1\frac{\partial h}{\partial \bm{q}}^T - \dot{\bm{q}_{p}}\right)\frac{1}{T_s},
		\label{eqn:avoid_fun_single}
	\end{align}
	where $k_1> 0$ is a constant that governs the strength related to joint position level avoidance.

	\subsubsection{Examples of avoidance acceleration formulation}\label{sec:avoid_acc_types}
	
	By introducing suitable functions for $h(\bm{q})$, the avoidance acceleration can be determined by using \eqref{eqn:avoid_fun_single}. Examples of different types of undesirable situations, their avoidance acceleration formulations, and how to combine them to determine the final avoidance acceleration are discussed below. 
	\paragraph{Loss of manipulability} The loss of manipulability $K(\bm{q})$ of a CDPR can be defined using \emph{unilateral dexterity} \citep{Kurtz_Hayward_94_DexterityMeasuresNPlusOneCables}
	\begin{align}
		K(\bm{q}) = \frac{\sigma_n(\bm{L}(\bm{q}))}{\sigma_1(\bm{L}(\bm{q}))}\frac{\sqrt{m+1}\hspace{0.5ex}\eta_{min}(\bm{q})}{\sqrt{\eta_{min}^2(\bm{q})+1}}. \label{eqn:uni_lat}
	\end{align}
	In \eqref{eqn:uni_lat}, $\sigma_1$ and $\sigma_n$ refers to the largest and smallest singular value of the joint-cable Jacobian matrix $\bm{L}(\bm{q})$, respectively, and $\eta_{min}(\bm{q})$ refers to the smallest element in the normalised null space vector $\hat{\bm{n}}(\bm{q})$, which is obtained by projecting the vector $\bm{v} = [1,1,...,1]^T \in \mathbb{R}^m$ into the null space of $\bm{L}(\bm{q})$
	\begin{align}
		\bm{n}(\bm{q}) = (\bm{I} - \bm{L}(\bm{q})\bm{L}^\dagger(\bm{q}))\bm{v}, \label{eqn:uni_lat_h}
	\end{align}
	where $\hat{\bm{n}}(\bm{q}) =  {\bm{n}(\bm{q})}/({\norm{\bm{n}(\bm{q})}})$. Consequently, $K(\bm{q})$ reaches a maximum value of $1$ when the system is \emph{unilateral isotropic} \citep{Kurtz_Hayward_94_DexterityMeasuresNPlusOneCables}, and reaches $0$ when the system loses its capability to generate any arbitrary direction of wrench through positive cable forces. Hence, the larger the $K(\bm{q})$, the higher the manipulability. Therefore, the avoidance of low manipulability acceleration $\ddot{\bm{q}}_{A_K}$ can be obtained by setting $h(\bm{q}) = K(\bm{q})$ in \eqref{eqn:avoid_fun_single}.
	
	\paragraph{Cable-link interference} For a CDPR, the distance between each \textit{cable-link pair} (SpiderArm) can be calculated numerically by treating cables and links as line segments \citep{Lumelsky_85_SegmentDistance}, denoted as $\bm{\delta}_{i}(\bm{q})$. Hence, the avoidance of interference acceleration, denoted by $\ddot{\bm{q}}_{A_D}$, can be generated by setting $h(\bm{q}) = \delta_{min}(\bm{q})$ in \eqref{eqn:avoid_fun_single}, where $\delta_{min}(\bm{q})$ is the distance between the closest cable-link pair in the CDPR.

	
	\paragraph{Joint limits}
	In MCDRs or HCDRs, hitting the mechanical joint limits is an undesirable situation, which causes instability in the system. The instability can be prevented by assuming
	\begin{align}
		h(\bm{q})= 
		\begin{cases}
			\frac{1}{2}(\bm{q}-\bm{q}_{max})^{2},& \bm{q}\ge\bm{q}_{max}\\
			\frac{1}{2}(\bm{q}-\bm{q}_{min})^{2}, &  \bm{q}\le\bm{q}_{min}\\
			0, & \text{otherwise}
		\end{cases}, \label{eqn:jt_lt}
	\end{align}
	where $\bm{q}_{max}$ and $\bm{q}_{min}$ represent the maximum and minimum joint limits, respectively. The avoidance of joint limits acceleration, denoted by $\ddot{\bm{q}}_{A_L}$, can be generated by setting $h(\bm{q})$ as per \eqref{eqn:jt_lt} in \eqref{eqn:avoid_fun_single}. 
	
	\paragraph{Combining multiple undesirable situations} In cases where more than one type of undesirable situations have to be avoided, ${\ddot{\bm{q}}}_A$ can be defined as the weighted average of $N$ joint accelerations
	\begin{align}
		{\ddot{\bm{q}}}_A = \sum_{i=1}^N w_i(\bm{q}, \dot{\bm{q}}) \cdot \ddot{\bm{q}}_{a,i},
		\label{eqn:avoid_fun_multi_weighted}
	\end{align}
	where $\ddot{\bm{q}}_{a,i}$ corresponds to the joint acceleration required to avoid the $i^{th}$ undesirable situations given by \eqref{eqn:avoid_fun_single}, and $w_i > 0$ represents the weights that priortize the avoidance of multiple situations. For example, if two types of undesirable situations are considered for a CDPR such as the loss of manipulability and the cable interference, then the avoiding acceleration ${\ddot{\bm{q}}}_A$ can be written as
	\begin{align}
		\begin{split}
			{\ddot{\bm{q}}}_A &= \ddot{\bm{q}}_{A_K} + w(\bm{q})\ddot{\bm{q}}_{A_D},\\
			w(\bm{q}) &={1}/\left(1+e^{\lambda_w({\delta_{min}{(\bm{q}}})-\epsilon)}\right),
		\end{split}\label{eqn:weight}
	\end{align}
	where $w(\bm{q})$ is a weight function that governs the acceleration that avoids cable-link interference $\ddot{\bm{q}}_{D}$. In \eqref{eqn:weight}, the cable-link pair minimum distance $ {\delta_{min}{(\bm{q}}})$ is determined by taking the minimum of $\{\bm{\delta}_{i}(\bm{q})\}_{i=1}^{n_l} $, where  $n_l$ represents the number of cable-link pairs. The weight function $ w(\bm{q})$ increases as the $ {\delta_{min}{(\bm{q}}})$ reaches the user-defined buffer $\epsilon$. The rate of $w(\bm{q}) $ increase is controlled by the value of $\lambda_w$. For a high value of $ \lambda_w $, $w(\bm{q}) $ approaches the value of unity, when $ {\delta_{min}{(\bm{q}}})$ just goes below the $\epsilon$. The weight of avoidance of low manipulability acceleration $\ddot{\bm{q}}_{K}$ is set to unity so that loss of manipulability avoidance can always be prioritized.
	
	\section{Iterative-Learning Controller}\label{sec:ilc}
	\begin{figure*}[htb!]
		\centering
		\def\svgwidth{\fs\textwidth}
		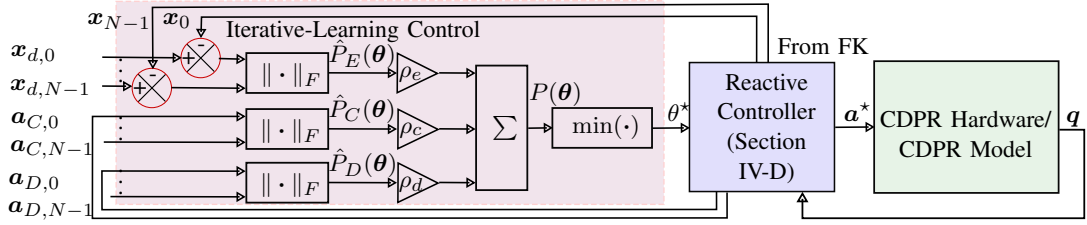
		\caption{ILC implementation details in the tri-space operational control framework.}
		\label{fig:osframework_ilc}
	\end{figure*}

	\ac{ILC} updates a set of control input trajectories when a system such as the CDPR is tracking a trajectory repeatedly over a finite time interval. For CDPRs, the control input would naturally be the cable forces $\bm{f}(t)\in\mathbb{R}^{m}$ and joint torques $ \bm{\tau}(t)\in\mathbb{R}^{p} $, combined to form $ \bm{a}(t)\in\mathbb{R}^{m+p} $. However, due to the cable force constraints and actuation redundancy, existing ILCs cannot be directly applied onto MCDRs or HCDRs to determine the cable forces. This is partially due to the lack of a fixed or iteration invariant relationship between the control inputs and system outputs coming from redundancy \citep{Tayebi_04_Adaptive_ILC,Janssens_Pipeleers_Swevers_13_Data_Driven_ILC_LTI}, as demonstrated in Section \ref{sec:Reactive}.
	In order to solve the redundancy problem, optimization techniques have been used by various  numerical gradient or search-based methods to find some optimal control input \citep{Kuc_Nam_Lee_91_ILC_Robot_Manipulators}. 
	However, given the high dimensions of the problem due to the high dimensional input signal $\mathbb{R}^{(m+p)\times N}$ for $m+p$ dimensional input and $N$ sampling points, it is not feasible to  achieve real-time control using ILC directly.
	
	This work proposes a novel Reactive-Iterative Tri-Space Operational Control Framework (see Section \ref{sec:OSFramework}), which  combines ILC with the reactive controller (see Section \ref{sec:Reactive}, Fig. \ref{fig:osframework0}). 
	Rather than iteratively learning the actuation command $ \bm{a}(t) $ in $\mathbb{R}^{(m+ p)\times N}$, it is proposed to learn a reactive tuning parameter vector $\btheta\in \mathbb{R}^{n+2}$ as discussed in  Section \ref{sec:OSFramework}. Note that the dimension of the parameter is much lower than the input vector, and hence the ``learned'' parameters are used within RC. 
	
	The reactive tuning parameter vector is constant for an entire trajectory and it is designed to affect the characteristic of the controller in resolving the joint to operational space redundancy (kinematic redundancy). Hence, the joint space profile for achieving better controller performance is explored by updating $\bm{\theta}$, while executing the desired operational space trajectory. Furthermore, the null space exploration parameters are unconstrained and, hence can be solved more easily. Mathematically, the ILC can be solved as an unconstrained nonlinear optimization problem
	\bea
	\btheta^\star = \argmin_{\btheta \in \mathbb{R}^{n+2}}P(\btheta),
	\label{eqn:ilc_nonlin_op_0}
	\eea
	where $P(\cdot)$ is the trajectory performance function, which will be defined later. The optimal set of parameter $\btheta^\star$ contains the the optimal set of null space parameters and the optimal $\alpha^*$ and $\beta^*$.

	As the unconstrained optimization problem \eqref{eqn:ilc_nonlin_op_0} is only performed once for each iteration or cycle off-line, there are many off-the shelf optimization techniques can be used. Practically, the parameter $\btheta$ is constrained in a known compact set $\Theta\in \mathbb{R}^{n+2}$. {It is assumed that such an optimal value exists as stated in the following assumption.}
	\begin{assumption}\label{assumption_compact set}
		{If the solution of the QP formulation} (\ref{eqn:reactive_qp}) {is feasible for any given initial $\btheta_0$, there exists a compact set $\Theta_C$ such that the optimal $\btheta^*$ exists uniquely in this compact set $\Theta_C$ satisfying $P(\btheta^*)=0$.}
	\end{assumption}
	
	An example of such a compact set is $\Theta_C:=\left\{\btheta\in  \mathbb{R}^{n+2}\left|\norm{\btheta}_2\leq C\right.\right\}$, for some positive constant $C$.
	This leads to the following optimization problem:
	\bea
	\btheta^\star = \argmin_{\btheta \in \Theta}P(\btheta).
	\label{eqn:ilc_nonlin_op}
	\eea
	It is usually assumed that this compact set is known. 
	\begin{remark}
		{
			It is to be noted that for a given task in the operational space and a given setting of CDPR, the optimal $\btheta^\star$ is fixed or iteration invariant. As the task is repetitive, the goal of ILC is to learn $\btheta^*$. In engineering applications, iteration-varying uncertainties and random noises exist. In this work, it is assumed that these iteration-varying uncertainties and noises are not dominant so that the task is almost repetitive. As shown in experimental results, the proposed framework can improve the performance over iterations.} \hfill $\circ$
	\end{remark}
	
	\subsection{Choices of $P(\cdot)$}\label{sec:traj_based_perf}
	The evaluation of the trajectory performance, such as tracking accuracy and actuation efforts, is done after each iteration. A few performance indices are considered. They are \emph{tracking accuracy} $P_{E}(\cdot)$, \textit{cable actuation effort} $P_{C}(\cdot)$, and \textit{direct actuation effort} $P_{D}(\cdot)$ (Fig. \ref{fig:osframework0} (b)). Other performance indices for the trajectories over iterations can be considered.
	
	Here ${P}_{E}$ represents the tracking performance, which is defined as
	\bea
	{P}_{E}(\btheta)=\norm{\bm{{E}}(\btheta)}_F,\label{P_E}
	\eea
	where the vector $\bm{E}$ is defined as
	\beas
	\bm{E}(\btheta) =[\bm{e}_{0}(\btheta) ,\cdots,\bm{e}_{N-1}(\btheta) ]\in\mathbb{R}^{r\times N}, \label{P_E}
	\eeas
	where $\bm{e}$ is defined in (\ref{error}). This performance $\bm{E}(\btheta)$ is a function of the tracking error $\bm{e}$. The cable actuation effort $P_{C}(\cdot)$ is defined as 
	\bea
	P_{C}(\btheta)=\norm{\bm{{A_C}}(\theta)}_F,\label{P_C}
	\eea
	where $\bm{A_C}(\btheta) = [\bm{a}_{C,0}(\btheta),\cdots,\bm{a}_{C,N-1}(\btheta)]\in\mathbb{R}^{m\times N}$ and $  \{\bm{a}_{C,k}\} \in\mathbb{R}^{m}$ are vectors of cable actuation effort at the $k^{th}$ sampling instants, $\forall k=0,\ldots, N-1$ (see Equation (\ref{eqn:a})). The direct actuation effort $P_{D}(\cdot)$ is defined as 
	\bea
	P_{D}(\btheta)=\norm{\bm{{A_D}}(\btheta)}_F,\label{P_D}
	\eea
	where $\bm{A_D}(\btheta) = [\bm{a}_{D,0}(\btheta),\cdots,\bm{a}_{D,N-1}(\btheta)]\in\mathbb{R}^{p\times N}$, and $ \bm{a}_{D,k} \in\mathbb{R}^{p}$ are vectors of direct actuation effort from active joint torque at the $k^{th}$ sampling instants, $\forall k=0,\ldots, N-1$ (see Equation (\ref{eqn:a})).

	\subsubsection*{Normalization}
	
	As different performance functions usually vary in scale and dimension, hence they are normalized first where,
	\bea
	P(\btheta) =\rho_e\hat{P}_{E}(\btheta)+\rho_c\hat{P}_{C}(\btheta)+\rho_d\hat{P}_D(\btheta),
	\label{eqn:ilc_P}
	\eea
	where $\rho_E$, $\rho_C$, and $\rho_D$ are positive normalization parameters. Here, $\hat P_E(\cdot)$, $\hat P_C(\cdot)$, and $\hat P_D(\cdot)$ are normalized tracking accuracy, normalized actuation effector and normalized direct actuation effort respectively. Usually, the normalization takes the following form, taking the example of $P_E(\cdot)$:
	\beas
	\hat P_E(\btheta)=\frac{P_E(\btheta)}{P_E^\star},
	\eeas
	where $P_E^\star=\displaystyle \min_{\btheta\in \Theta}P_E(\btheta)$ as $P(\cdot)$ is an unknown nonlinear mapping. Here $\Theta$ is the set containing the all computed parameters so far. Similarly, $P_C^\star=\displaystyle \min_{\btheta\in \Theta}P_C(\btheta)$ and $P_D^\star=\displaystyle \min_{\btheta\in \Theta}P_D(\btheta)$. These unknown parameters or values can be learnt from the ILC algorithm.
	
	In contrast to the reactive null-space projection methods that aim to optimize certain objective function at each time step \citep{DeSapio_Khatib_Delp_08_LeastAction,Khatib_87_UnifiedOpSpaceFormulation}, which requires to compute the optimal ${\bm a}\in \mathbb{R}^{(m+p)\times N}$, the proposed ILC tries to update the null space vector $ \bm{\ddot{{ q}}}_{c_N} $ (see Equation \eqref{eqn:q_c_N}) at each iterations (slower time-scale) by updating  the set of null space exploration parameters $\btheta \in \mathbb{R}^{n+2}$, defined by \eqref{theta}. The objective of the ILC algorithm is to identify the optimal joint space behavior, i.e., find $P_E^\star$, $P_C^\star$, $P_D^\star$, and $\btheta^{\star}$. The ILC implementation details are shown in Fig. \ref{fig:osframework_ilc}.
	
	\subsection{The Choice of Optimization Methods}\label{sec:opt_method}
	
	Although the null space exploration parameters provide a great potential in exploiting the kinematic redundancy and improving in task performance, the highly nonlinear  relationship between these parameters and the trajectory performance function $P(\cdot)$ is unknown, making it hard to find an optimal solution without knowing the explicit gradient information. Thus the standard gradient-based optimization methods such as Broyden–Fletcher–Goldfarb–Shanno (BFGS) algorithm \citep{Liu_Nocedal_89_LBFGS} not applicable for solving \eqref{eqn:ilc_nonlin_op} and \eqref{eqn:ilc_P}.
	
	Hence, in this work, search-based methods such as \textit{\ac{PS}} \citep{Hooke_Jeeves_61_DirectSearch} and {\ac{PSO}} \citep{Kennedy_Eberhart_95_ParticleSwarmOptimization} are applied to determine $\btheta^\star$ within a given compact set $\Theta_C$. Both techniques can find an optimal solution without the gradient information. 
	
	In simulations, PS was used to as the Algorithm \ref{alg:ilc_ps} to ensure its convergence due to introduction of convergence parameter $\rho_{ILC}$. On the other hand, PSO is used in the experiments since \ac{PSO} is well-known for its ability in terms of robustness with respect to measurement noises, though the convergence of PSO is hard to show. Besides, PSO is used for the hardware mode as the hardware uncertainties (such as friction and sensor noise) affects the performance evaluation for the same set of control parameters. 
	
	Next, the PS technique is used to iteratively solve the optimization problems \eqref{eqn:ilc_nonlin_op} and \eqref{eqn:ilc_P} as shown in Algorithm \ref{alg:ilc_ps}. It is known that finding $P(\btheta^\star)$ is similar to find $P_E^\star$, $P_C^\star$, and $P_D^\star$. Algorithm 1 is run three times to solve $Q_E^\star$, $Q_C^\star$, and $Q_D^\star$. Then, Algorithm 1 is run to solve the optimization problem of \eqref{eqn:ilc_P}, when these values are computed. It is observed during simulations that $P_E^\star$, $P_C^\star$, and $P_D^\star$ are sensitive to the desired trajectory $\bm{x}_d$ or $\mathcal{X}_d$, hence, they are computed once separately for a given $\mathcal{X}_d$. 
	
	\begin{algorithm}[htb!]
		\caption{Proposed Control framework using \acf{PS} for ILC}
		\label{alg:ilc_ps}
		\begin{algorithmic}[1]
			\Require Reference traj $\mathcal{X}_d=\{\bm{x}_{d}(t), \dot{\bm{x}}_{d}(t), \ddot{\bm{x}}_{d}(t)\}\in\mathbb{R}^{r}$
			\Require Initial null space exploration parameters $\btheta_0\in\mathbb{R}^{n+2}$
			\Require Step size $\delta_\theta\in\mathbb{R}$
			\Require The bound of $\btheta$ (the compact set $\Theta$)
			\Require Convergence parameter $\rho_{ILC}\in (0,1)$
			\Ensure Optimal null space exploration parameters $\btheta^\star\in\mathbb{R}^{n+2}$
			\Ensure  Optimal joint space position trajectory $\bm{Q}^\star\in\mathbb{R}^{n\times N}$
			\Ensure Optimal actuation commands traj $\bm{A}^\star\in\mathbb{R}^{(m+p)\times N}$
			\State $P^\star \in\mathbb{R}\gets \infty$ (initialise $P^\star$) \label{alg:ps_init}
			\For{$i$-th iteration of ILC}
			\State // \text{Explore neighborhood of $\btheta_i=\left[\theta_{i,1},\cdots,\theta_{i,n+2}\right]^T$} 
			\For {its  $j^{th}$ element $\theta_{i,j}$}
			\State $\overline{\theta}_{i,j} \gets \theta_{i,j}$ + $\delta_\theta$ (update of $\btheta$) \label{alg:ps_u}
			\State // Execute trajectory with updated $\overline{\btheta}_{i}$
			\State Run Algorithm \ref{alg:rqp} with $\mathcal{X}_d$, $\overline{\btheta}_{i}$ to obtain $\bm{Q}, \bm{A}$  \label{alg:ps_P_start}
			\State $P(\overline{\btheta}_{i}) \gets$ \eqref{eqn:ilc_P} using $\bm{Q}(\overline{\btheta}_{i})$ and $\bm{A}(\overline{\btheta}_{i})$
			\EndFor
			\State $P^\star_i \gets$ best $P$ in the $i$-th iteration
			\State $\btheta^\star_i \gets$ corresponding $\overline{\btheta}_{i}$ to $P^\star_i$ \label{alg:ps_uistar}
			\If{$P^\star_i \leq \rho_{ILC}  P^\star$ and $\btheta_i\in \Theta$}
			\State $\btheta^\star \gets \btheta^\star_i$;\label{alg:ps_ustar}
			\State $\bm{Q}^\star \gets \bm{Q}(\btheta^\star)$;\label{alg:ps_Qstar}
			\State $\bm{A}^\star \gets \bm{A}(\btheta^\star)$;\label{alg:ps_Astar}
			\Else
			\State $\delta_\theta \gets \frac{\delta_\theta}{2}$
			\State $\btheta_i\gets Proj(\btheta_i,\Theta)$
			
			\EndIf
			\EndFor
			\State \Return $\btheta^\star, \bm{Q}^\star, \bm{A}^\star$
		\end{algorithmic}
	\end{algorithm}
	
	\begin{algorithm}[htb!]
		\caption{Reactive Controller for a single trajectory}
		\label{alg:rqp}
		\begin{algorithmic}[1]
			\Require Reference traj $\mathcal{X}_d=\{\bm{x}_{d}(t), \dot{\bm{x}}_{d}(t), \ddot{\bm{x}}_{d}(t)\}\in\mathbb{R}^{3r}$
			\Require  Null space exploration parameters $\btheta_0\in\mathbb{R}^{n+2}$
			\Ensure Joint space position trajectory $\bm{Q}$
			\Ensure Actuation commands trajectory $\bm{A}$
			\For{$k \gets 0, \dots, N-1$}
			\State $\bm{q}_i, \dot{\bm{q}}_i  \gets$ from robot state
			\State $g_t(\mathcal{X}_d, \ddot{\bm{q}}, \btheta) \gets$ \eqref{eqn:g_t}
			\State $g_f \gets$ \eqref{eqn:rqp_actuation_obj}
			\State $g_a \gets$ \eqref{eqn:avoid_fun_general} - \eqref{eqn:avoid_fun_multi_weighted}
			\State $\bm{\Phi}, \bm{\rho} \gets$ \eqref{eqn:react_hardcon_q_dd}
			\State $\ddot{\bm{q}}_i, \bm{a}_i \gets$ solving \eqref{eqn:reactive_qp} using $\bm{q}_i, \dot{\bm{q}}_i, g_t, g_f, g_a, \bm{\Phi}, \bm{\rho}$
			\State Use $\ddot{\bm{q}}_i, \bm{a}_i$ to execute on the robot
			\State Add $\bm{q}_i$ to $\bm{Q}$
			\State Add $\bm{a}_i$ to $\bm{A}$
			\EndFor
			\State \Return $\bm{Q}, \bm{A}$
		\end{algorithmic}
	\end{algorithm}
	
	Here, $Proj(\btheta,\Theta)$ constrains the updating law of $\btheta_i$ within the compact set $\Theta$. There are many different ways to select the compact set $\Theta$. One of them is to choose $\Theta:=\left\{\left.\theta_j\right|\underline{\theta}_j\leq\theta_j\leq \bar \theta_j, j=1,\ldots, n+2\right\}$ so that 
	\beas
	Proj\left( \theta_j,(\underline{\theta}_j, \bar{\theta}_j)\right)=\left\{\begin{array}{cc} \underline{\theta}_j&\theta\leq \underline{\theta}_j\\ 
		\theta_j& \underline{\theta}_j\leq \theta_j\leq \bar{\theta}_j\\
		\bar{\theta}_j&\theta_j\geq \bar{\theta}_j\end{array}\right.,
	\eeas
	which can ensure that $\btheta_i\in \Theta_C$. At each iteration, new 'optimal' value $\btheta_{i+1}=\btheta^\star= \min_{s=0,\ldots, i+1}\left\{P(\btheta_s)\right\}$ is computed. Algorithm \ref{alg:ilc_ps} will ensure that at the $(i+1)^{th}$ iteration, the performance function will improve, i.e., $P(\btheta_{i+1})\leq \rho_{ILC} P(\btheta_i)$ for some $\rho_{ILC}\in (0.1,1)$, leading to the convergence of parameter $\btheta$ to the optimal value $\btheta^\star$. Moreover, the constraint set $\Theta_C$ is used in Algorithm \ref{alg:ilc_ps} will ensure the boundedness of the parameter updating. 
	
	Once this new optimal value is computed, it will serve as an input to the \ac{RC} (Fig. \ref{fig:osframework0}), which uses a \ac{QP} to solve the optimization problem (Equation \eqref{eqn:reactive_qp}) as summarized in Algorithm \ref{alg:rqp}. Algorithm \ref{alg:ilc_ps} and \ref{alg:rqp} as well as Fig. \ref{fig:osframework0} and \ref{fig:osframework_ilc} also demonstrate the interaction between the ILC and the \ac{RC}. With the generalizabilty of the framework, other optimization methods, such as the \ac{PSO}, can be easily implemented by changing the initialisation procedure and the update method of $\btheta$ (Lines \ref{alg:ps_init} and \ref{alg:ps_u}, Algorithm \ref{alg:ilc_ps}, respectively). Algorithm \ref{alg:rqp} presents the reactive controller for a single trajectory. For the convenience of notations, the set of joint space position and actuation command trajectories  are denoted as $\bm{Q}$ and $\bm{A}$, respectively. 
	
	\section{Stability Analysis}\label{section:stability}
	
	The stability of the class of controllers that minimizes the actuation effort given by \eqref{eqn:rqp_actuation_obj} is a well-studied topic \citep{Peter_08_LearningControlOperationalSpace, Hsu_89_DynamicControlRedundantManipulators} for unconstrainted cases. In particular, the gradient of the \ac{QP} problem (\ref{eqn:reactive_qp}) without constraints is computed as the reactive control law $\bm{a}$ to show the stability.  However, due to the existence of constraints, such an analysis is not sufficient to show the boundedness of the trajectories.

	In this work perturbation theory \cite[Chapter 10]{Khalil_2002_Nonlinear} is applied for the stability analysis. The analysis shows the boundedness of trajectories for any sampling instant $k=0,\ldots, N-1$ and for any iteration, when both the \ac{RC} law in \eqref{eqn:reactive_qp} (see Algorithm \ref{alg:rqp}) and \ac{ILC} law (see Algorithm \ref{alg:ilc_ps}) are working together. Then the ILC algorithm (Algorithm \ref{alg:ilc_ps}) can ensure the tracking convergence. Hence, the algorithm's convergence guarantees tracking performance improvement, which can be seen in Table \ref{tab:perf_eval}.   
	
	\subsection{Boundedness of Trajectories} 
	
	At the $i^{th}$ iteration,  any computed $\btheta_i$ from Algorithm \ref{alg:ilc_ps} is constrained in the compact set $\Theta$.  Consequently, it leads to the bounded solutions of QP problem  $\bm{a}_k^i$ and ${\ddot {\bm{q}}}_k^i$ (Equation \eqref{eqn:reactive_qp} and Algorithm \ref{alg:rqp}). By applying perturbation theory \citep[Chapter 10]{Khalil_2002_Nonlinear} for a finite interval, the boundedness of ${\ddot {\bm{q}}}_k^i$ leads to the boundedness of ${\dot {\bm{q}}}_k^i$ and $\bm{q}_k^i$. 
	Hence, at the $i^{th}$ iteration and within a fixed sampling time $T_s$, the sampled-data structure of actuation $\bm{a}^i (t)$ in \eqref{eqn:eom_general} becomes
	\begin{align}
		\bm{M}(\bm{q}^i(t))\ddot{\bm{q}}^i(t)&+ \bm{C}(\dot{\bm{q}}^i(t),\bm{q}^i(t)) + \bm{G}(\bm{q}^i(t)) = \bm{B}(\bm{q}^i(t))\bm{a}^i(t),  \nonumber\\
		\text{where, }\bm{a}^i(t) &= \bm{a}_k^i, \forall t\in [kT_s, (k+1)T_s),  \label{sampled-data}
	\end{align}
	for all $k=0,\ldots, N-1$. In \eqref{sampled-data}, $\bm{a}_k^i$ is computed from Algorithms \ref{alg:ilc_ps} and \ref{alg:rqp}. The continuity of the trajectories from  (\ref{eqn:eom_general}) ensures that $\bm{a}^i(t)$, $\bm{q}^i(t)$, ${\dot {\bm{q}}}^i(t)$, and ${\ddot {\bm{q}}}^i(t)$ are uniformly bounded for any $t\in [0,T]$ for the $i^{th}$ iteration.  This leads to the following theorem.
	\begin{theorem}
		\emph{Assume that  a CDPR system  \eqref{eqn:eom_general} has a sampled-data structure \eqref{sampled-data} with constraints \eqref{eqn:eom_general} to \eqref{constraints}. If actuation $\bm{a}_k^i$ is  computed from Algorithms \ref{alg:ilc_ps} and \ref{alg:rqp}, then  $\bm{q}^i(t)$, ${\dot {\bm{q}}}^i(t)$, and ${\ddot {\bm{q}}}^i(t)$ in \eqref{sampled-data} are uniformly bounded for any $t\in [0,T]$ for any $i^{th}$ iteration. Moreover constraints \eqref{constraints} are satisfied from QP (Algorithm \ref{alg:rqp}).}  \hfill $\circ$
	\end{theorem}
	\begin{remark}
		{It is noted that a Proportional-Derivative (PD) controller is used in the designed framework. By tuning the gains of PD controller, it is possible to tune the tracking error at the $i^{th}$ iteration, the $k^{th}$ sampling instant $\bm{e}_k^i, i=0,\ldots;k=0,\ldots, N-1$ to an appropriate range. This indicates that $\bm{x}^i(t)$ is uniformly bounded for any iteration and any $t\in [0,T]$. By using pseudoinverse $J$, we can also conclude that $\bm{q}(t)$ is also bounded. Such an analysis as in \citep{Peter_08_LearningControlOperationalSpace, Hsu_89_DynamicControlRedundantManipulators} can provide less conservative estimation of the bound of trajectories. As QP \eqref{eqn:rqp_actuation_obj} already provided bounded solutions, the presented analysis is simpler. It is highlighted that the performance improvement along the iteration domain is achieved by Algorithm \ref{alg:ilc_ps}. The role of \ac{RC} is to avoid unwanted behaviours in time domain $t\in [0,T]$, which is achieved by QP (Algorithm \ref{alg:rqp}).} \hfill $\circ$
	\end{remark}
	\begin{table*}[t]
		\caption{Control and trajectory parameters used for simulations.}
		\label{tab:control_param}
		\centering
		\resizebox{0.90\textwidth}{!}{\begin{tabular}{lccccccccccc}
				\toprule[0.8mm]
				\textbf{Row}&\textbf{CDPR}&\textbf{Fig.} & \multicolumn{1}{c}{$\alpha$}                & $ \beta $        & $ \bm{K}_p $     & $ \bm{K}_d $     & \textbf{Initial Joint Position} $\bm{q}_0 $         & \textbf{Initial Task Position} $\bm{x}_0 $         & \textbf{Frequency} (Hz) &\textbf{Result} \\\midrule
				1&\multirow{3}{*}{BMArm}&\ref{fig:bmarm_simulation_1} (b) to (d)    & $ 1\times10^{-6}  $    & 0              & $ 200\bm{I}_{3} $& $ 28\bm{I}_{3} $ & $ [0.18,0,0,-0.45]^{T} $ & $ [0,0.595,0]^{T} $ &100 &Success\\
				2&&\ref{fig:bmarm_simulation_1} (e) to (f), and \ref{fig:bmarm_simulation_2} (a)    & $ 1\times10^{-6}  $    & 0              & $ 200\bm{I}_{3} $& $ 28\bm{I}_{3} $ & $ [0.18,0.2,0,-0.45]^{T} $ & $ [0,0.595,0]^{T} $ &100 &Failure\\
				3&&\ref{fig:bmarm_simulation_2} (a) to (c) & $ 1\times10^{-6}  $    & $ 1\times10^{-2 }$ & $ 200\bm{I}_{3} $& $ 28\bm{I}_{3} $ & $ [0.18,0.2,0,-0.45]^{T} $ & $ [0,0.595,0]^{T} $ &100 &Success\\
				\midrule
				4&\multirow{1}{*}{SpiderArm}&\ref{fig:spiderarm_simulation} (b) to (l)     & $ 1\times10^{-6}  $    & $ 2\times10^{-1 }$ & \begin{tabular}[c]{@{}c@{}}Position: $ 40\bm{I}_{3}$\\  Orientation: $ 5\bm{I}_{3} $\end{tabular}&  \begin{tabular}[c]{@{}c@{}}Position: $ 12.6\bm{I}_{3}$\\  Orientation: $ 4.5\bm{I}_{3} $\end{tabular}& $ [0,0,-\pi/2]^{T} $ & $ [2,1.1,1.14]^{T} $ &100, 200&Success\\
				\midrule
				5&\multirow{1}{*}{FASTKIT}& \ref{fig:fastkit_simulation} (b) to (f)     & $ 1\times10^{-6}  $    & $ 1\times10^{-1 }$ &$200\bm{I}_{3} $& $ 20\bm{I}_{3} $ & $ [0,0,-\pi/2]^{T} $ & $ [1.5,0,1.35]^{T} $ &200&Success\\
				\bottomrule[0.8mm]
				
			\end{tabular}
		}
	\end{table*}
	\subsection{Convergence of ILC Algorithm}
	
	By introducing convergence parameter $\rho_{ILC}$ and the projection operator $Proj(\btheta,\Theta)$, the proposed \ac{PS} based ILC Algorithm \ref{alg:ilc_ps} ensures that the cost function $P(\cdot)$ decreases monotonically. If Assumption \ref{assumption_compact set} holds, Algorithm  \ref{alg:ilc_ps} can ensure that $\displaystyle \lim_{i\rightarrow \infty}P(\btheta_i^\star)=0$ and $\displaystyle \lim_{i\rightarrow \infty}\btheta_i^\star=\btheta^\star$, indicating that $\displaystyle \lim_{i\rightarrow \infty}E(\btheta_i^\star)=0$, which comes from (\ref{P_E}) and (\ref{eqn:ilc_P}). This shows that $\displaystyle \lim_{i\rightarrow \infty}\bm{e}_k^i(t)={\bf 0}, \forall k=0,\ldots, N-1$. 
	
	For any given $\delta>0$, by applying \citep[Theorem 1]{Nevsic_1999_Sufficient}, it can be shown that  $\displaystyle \limsup_{i\rightarrow \infty}\norm{\bm{e}^i(t)}_2\leq \delta, \forall t\in [0,T]$, by selecting a sufficiently small sampling period $T_s$. This is summarized in the following theorem.
	
	\begin{theorem}\label{t2}
		\emph{Assume that Assumption \ref{assumption_compact set} holds for some known compact set $\Theta_c$. Let $\delta$ be any positive constant. There exists $T_s^\star>0$ such that for any small sampling interval satisfying $T_s\in (0, T_s^\star)$,  the proposed Algorithm  \ref{alg:ilc_ps} can ensure that   $\displaystyle \limsup_{i\rightarrow \infty}\norm{\bm{e}^i(t)}_2\leq \delta, \forall t\in [0,T]$.} \hfill $\circ$
	\end{theorem}
	
	\begin{remark}
		{Theorem \ref{t2} shows that if Assumption \ref{assumption_compact set} holds, by using Algorithm \ref{alg:ilc_ps}, the operational space tracking error $\bm{e}^i(t)$  can be arbitrarily small as iteration number approaches to infinity. Consequently, Theorem \ref{t2} also suggested that for a given positive constant $\delta_1>\delta$, where $\delta$ comes from Theorem \ref{t2}, there exists $T_s^\star>0$ and $N^\star\in \mathbb{N}$ such that}
		\beas
		\displaystyle \norm{\bm{e}^i(t)}_2\leq \delta, \forall t\in [0,T], i\in \mathbb{N}  \  \mbox{and} \  i \geq N^\star, 
		\eeas
		to ensure the convergence in a finite number of iterations. \hfill $\circ$
	\end{remark}
	\begin{remark}
		{If there exists $\btheta^\star$ such that $P(\btheta^\star)\neq 0$, Algorithm \ref{alg:ilc_ps} can select $\rho_{ILC}=1$ to show that $\displaystyle \limsup_{i\rightarrow \infty} P(\btheta_i)=C_1$ for some positive constant $C_1$. The tracking error at steady-state in iteration $\displaystyle \limsup_{i\rightarrow \infty}\bm{e}^i(t)$ is still bounded. Such a bound is determined by the parameters $\rho_e$, $\rho_c$, and $\rho_d$ in \eqref{eqn:ilc_P}  as well as the value of $C_1$.} \hfill $\circ$
	\end{remark}
	
	\section{Simulation Results}\label{sec:sim_results}
	
	Simulation results on one \ac{MCDR}: BMArm, and two \acp{HCDR}: SpiderArm and FASTKIT are presented in this section to demonstrate the capability and features of the proposed framework. 
	
	Additionally, the simulation results also signify the ability of the proposed reactive controller to be applied to different systems with no controller tuning required. Table \ref{tab:control_param} provides all the parameters associated with the simulations. Frequencies $ 100 $ and $ 200 $ Hz were selected to evaluate the controller performance with different frequencies. 
	
	Controller gains $\bm{K}_p $, and $\bm{K}_d $ were selected in such a manner so that the damping ratio was approximately set to unity in all the cases and the boundedness of $\bm{e}(t), t\in [0,T]$ were achieved. For the \ac{ILC} optimization in the simulations, \ac{PS} was used as discussed in Section \ref{sec:opt_method}. All simulations were performed with the open-source \ac{CASPR} \citep{Lau_Eden_Tan_Oetomo_16_CASPR} software using a computer with an Intel Core i9-11900K CPU @ 3.50 GHz and 32.0 GB of RAM,  using MATLAB R2018a (64-bit). The QP in the reactive controller was solved using \textit{qpOASES} \citep{Ferreau_Kirches_Potschka_Bock_Diehl_14_qpOASES}.
	
	\subsection{BMArm}\label{sec:react_bmarm}
	
\begin{figure*}[!tb]
		\centering
		\includegraphics[width=\fss\textwidth]{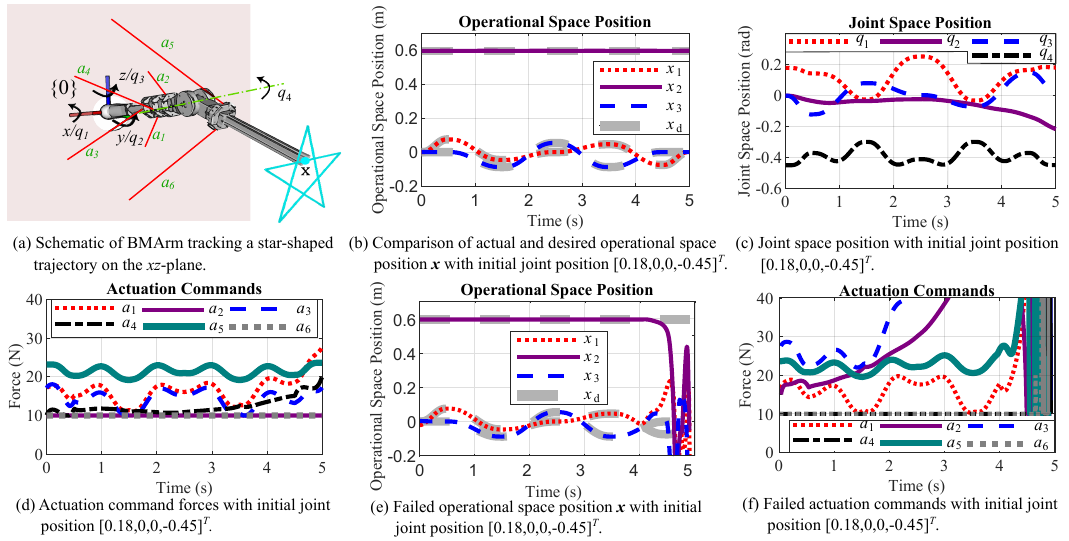}
		\caption{BMArm operational space tracking simulation results for star-shaped trajectory, without avoidance function  i.e $ \beta=0 $.}
		\label{fig:bmarm_simulation_1}
	\end{figure*}

	The BMArm (Fig. \ref{fig:bmarm_simulation_1} (a)) is a 4-\ac{DoF} \ac{MCDR} ($n = 4$). It consists of one spherical joint and one revolute joint. The joint space $\bm{q} \in \mathbb{R}^4$ is composed of $q_1, q_2$, $q_3$, and $q_4$, where $q_1, q_2$ and $q_3$ are the $xyz$-Euler angles of the spherical joint and $q_4$ is the angle of the revolute joint. The joint space is actuated by 6 cable forces $\bm{f} \in \mathbb{R}^6$, which forms the actuation command $ \bm{a}\in \mathbb{R}^6$ ($ \bm{a}=\bm{f} $). The operational space $\bm{x} \in \mathbb{R}^3$ is defined as the \emph{xyz}-coordinate of the second link's tip in frame \{0\}.
	
	\begin{figure*}[htb!]
		\centering
		\includegraphics[width=\fss\textwidth]{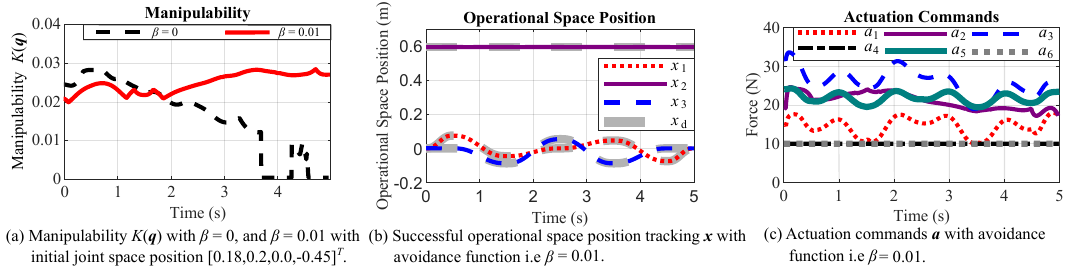}
		\caption{BMArm operational space tracking simulation results for star-shaped trajectory, with avoidance function i.e $ \beta=0.01 $.}
		\label{fig:bmarm_simulation_2}
	\end{figure*}
	To demonstrate the capability of resolving the kinematic and actuation redundancies with the tracking function $g_t(\mathcal{X}_d, \ddot{\bm{q}})$, actuation function $g_f(\bm{a})$, and avoidance function $g_a(\bm{\dot{q}})$, simulations on the BMArm are performed. Additionally, to show the relevance of the avoidance function, the simulations are performed in two parts: \textit{with and without avoidance function}.
	
	\subsubsection{\ac{RC} without avoidance function} A star-shaped trajectory on the \emph{xz}-plane (Fig. \ref{fig:bmarm_simulation_1} (a)) is tracked with control and trajectory parameters defined in Row 1, Table \ref{tab:control_param} with initial joint position is set at $\bm{q} = [0.18,0,0,-0.45]^T$. Fig. \ref{fig:bmarm_simulation_1} (b) shows that the reactive controller tracks the reference operational space position $\bm{x}_d$ while resolving the kinematic and actuation redundancies. The resulting joint space motion $\bm{q}$ and actuation commands $\bm{a}$ is shown in Fig. \ref{fig:bmarm_simulation_1} (c) and (d), respectively.
	
	Next, consider the simulation of the same star trajectory, but with a different initial joint position $\bm{q}=[0.18,0.2,0,-0.45]^T$ in which a slight twist of the system about the $y$-axis has been added (Row 2, Table \ref{tab:control_param}). The results in Fig. \ref{fig:bmarm_simulation_1} (e) to (f) show that the operational space tracking becomes unstable and eventually diverges. The manipulability metric $K(\bm{q})$, given by \eqref{eqn:uni_lat} (black dashed line, Fig. \ref{fig:bmarm_simulation_2} (a)) reaches $0$, which means that the system loses its capability to generate any direction of wrench and leads to the divergence behavior of the controller.
	
	\subsubsection{\ac{RC} with avoidance function} The avoidance acceleration was formulated by combining the avoidance of low manipulability and joint limits acceleration in the form of \eqref{eqn:weight}. Fig. \ref{fig:bmarm_simulation_2} demonstrates the effect of the avoidance of low manipulability by setting $h(\bm{q}) = K(\bm{q})$ for the same task as in Fig. \ref{fig:bmarm_simulation_1} (e) to (f), with parameters defined in Row 3, Table \ref{tab:control_param}. It can be seen from Fig. \ref{fig:bmarm_simulation_2} (b) and (c) that the tracking task is completed without divergence and significant improvement in manipulability  (red line, Fig. \ref{fig:bmarm_simulation_2} (a)) has been achieved as compared to Fig. \ref{fig:bmarm_simulation_2} (a), which demonstrates the ability of the avoidance function to maintain the system's manipulability to complete the tracking task. 
	
	\begin{figure*}[!h]
		\centering
		\includegraphics[width=\fss\textwidth]{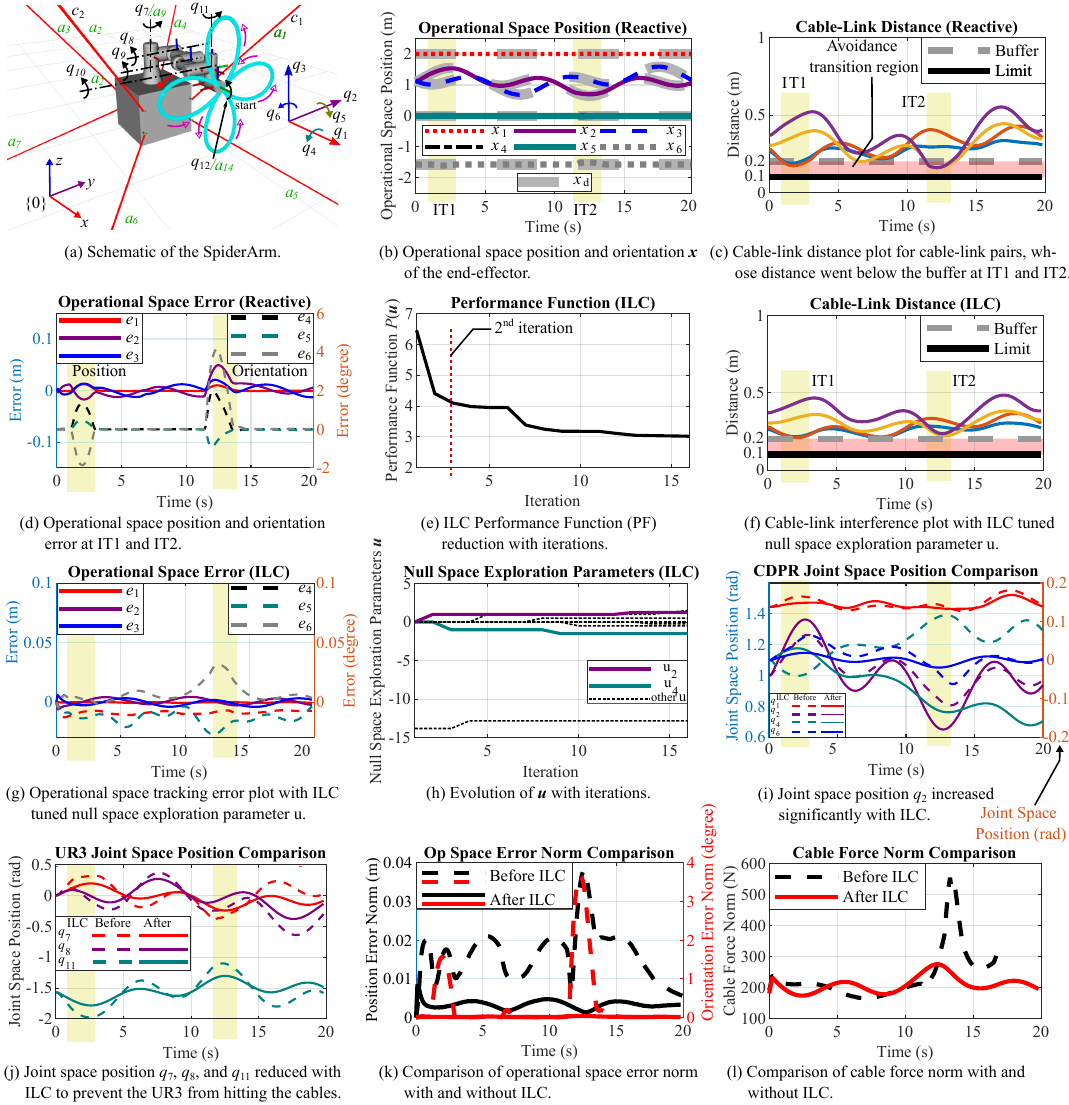}
		\caption{SpiderArm operational space tracking simulation results for flower-shaped trajectory.}
		\label{fig:spiderarm_simulation}
	\end{figure*}
	\subsection{SpiderArm}\label{section:spiderarm_simulation_results}
	
	The SpiderArm is an HCDR that combines a 6-\ac{DoF} UR3 manipulator on the end-effector of a spatial SCDR (Fig. \ref{fig:spiderarm_simulation} (a)). Additionally, the large translational workspace of SCDR, and the installation of the robot arm leads to increased dexterity. The hybrid robot is actuated by both cables and revolute joints, forming actuation command $\bm{a} \in \mathbb{R}^{14}$, where the SCDR is actuated by 8 cables $[a_1, a_2, \ldots, a_8]^T$ that are attached to the base frame, and the 6-\ac{DoF} robot arm is actuated by revolute joints $[a_9, a_{10}, \ldots, a_{14}]^T$. The joint space of the SpiderArm is defined as $\bm{q} = [q_1, q_2, \ldots, q_{12}]^T$, where $q_1, q_2$ and $q_3$ refer to the translation, $q_4, q_5$ and $q_6$ refer to the Euler orientation of the SCDR in frame $\{0\}$, and the joints $q_7, q_8, \ldots, q_{12}$ refer to the joint angles of the robot arm. The operational space is defined as $\bm{x} = [x_1, x_2, \ldots, x_6]^T$, which represents the \emph{xyz}-coordinate and Euler angles of the tip of the robot arm in frame $\{0\}$.  Simulations on the SpiderArm were performed to discuss the performance of RC, ILC and the proposed framework.
	
	\subsubsection{\acf{RC}}\label{section:spiderarm_simulation_results_reactive}
	A rose-shaped trajectory (Fig. \ref{fig:spiderarm_simulation} (a)) was tracked (Fig. \ref{fig:spiderarm_simulation} (b)) with parameters defined in Row 5, Table \ref{tab:control_param} for $T=20$ s to demonstrate the capability of the proposed reactive controller with avoidance acceleration. 
	
	The possibility of the SpiderArm (Fig. \ref{fig:spiderarm_simulation} (a)) hitting the cables due to the large amplitude of the trajectory ($0.45$m), influenced the consideration of two types of undesirable situations: loss of manipulability and cable-link interference (see Section \ref{sec:react_avoid}), which were combined as per \eqref{eqn:weight}. The buffer $\epsilon$ in \eqref{eqn:weight} was set at $0.2$ m (dashed line in Fig. \ref{fig:spiderarm_simulation} (c)), and the hard limit was set at $0.1$ m (solid horizontal line in Fig. \ref{fig:spiderarm_simulation} (c)), which stopped the system once encountered (see Section \ref{sec:avoid_acc_types} b)). The cable-link avoidance acceleration was obtained by setting $h(\bm{q})$ to $\delta_{min}(\bm{q})$, where $\delta_{min}(\bm{q})$ refers to the minimum distance between the UR3 manipulator link and the closest cable. 
	
	In SpiderArm, $n_l= 48$ cable-link pairs were considered such that when one of the distance crosses below the  $\epsilon$ and enters the \textit{avoidance transition region} (red shaded region in Fig. \ref{fig:spiderarm_simulation} (c)), the weight function $ w(\bm{q}) $ in \eqref{eqn:weight} reaches unity abruptly, and avoidance acceleration for cable-link interference gets more weightage. For low value of $ w(\bm{q}) $, the controller focused on maintaining the manipulability of the HCDR. Since the framework is generic so other types of interference such as cable-cable interference can also be incorporated in a similar manner in the form of \eqref{eqn:weight}.  
	
	Fig. \ref{fig:spiderarm_simulation} (c) illustrates those cable-link pairs, which crosses the buffer $ \epsilon $ during the tracking task. At Interference Time IT1 and IT2 (highlighted by yellow), the distance between a cable-link pair dropped below the buffer of $0.2$ m (grey dashed line), and hence the system entered the avoidance transition region. With the avoidance function for cable interference, the controller increased the distance of the worst-case cable-link pair, and pushed the curve above the avoidance transition region. However, due to the weighted formulation of the \ac{QP}, the need for avoidance led to reduced emphasis on tracking, leading to higher tracking error at IT1, and IT2 (Fig. \ref{fig:spiderarm_simulation} (d)). 
	
	\subsubsection{\acf{ILC}}\label{section:spiderarm_simulation_results_ilc} To improve the tracking performance of RC, the null space of the system was explored by the proposed ILC by optimizing the trajectory performance function $P(\btheta)$ described in \eqref{eqn:ilc_P}, which consists of the weighted sum of the normalized tracking error norm, and normalized actuation effort norm made by the cables and the joints of the CDPR and the UR3, respectively (see Section \ref{sec:traj_based_perf}). The weights chosen for the final performance function components were $\rho_E=\rho_C=\rho_D=1  $. $P(\btheta)$ was optimized by the \ac{PS} algorithm (Algorithm \ref{alg:ilc_ps}) with initial null space exploration parameters $\btheta$ and step size set at $[0,...0,\ln(1\times10^{-6}), \ln(1\times10^{-2})]^T$,  and unity, respectively. For $\btheta \in \mathbb{R}^{14}$, the \ac{PS} algorithm has performed $14\times2+1=29$ evaluations in each iteration. Hence, for a total of $15$ iterations, $435$ $(29\times15)$ evaluations were performed.
	
	After the optimization, $P(\theta)$ reduced significantly within the time interval of $6.5$ to $7$ s (Fig. \ref{fig:spiderarm_simulation} (e)) since the ILC discovered a joint space trajectory that does not cause cable interference. From Fig. \ref{fig:spiderarm_simulation} (f), the distances between cable-link pairs stayed above the buffer (grey dashed) throughout the entire trajectory. As a result, the avoidance function was not activated during the task, and the RC emphasized more on the tracking task. Hence, the tracking error maintained at a low level (Fig. \ref{fig:spiderarm_simulation} (g)) and resulted in a low value of $P(\btheta)$, which demonstrates that the ILC has the potential to produce a motion that both decreases operational space error, and it can avoid interference between the robot arm and the cables simultaneously, by exploiting the joint space redundancy throughout the trajectory.

	Additionally, further insights regarding the task were also obtained from the null space exploration parameters. It can be observed from Fig. \ref{fig:spiderarm_simulation} (e) that the performance function experienced the most significant decrease in the first $2$ iterations. From Fig. \ref{fig:spiderarm_simulation} (h), it can be seen that the null space exploration parameter $ u_2 $ (solid purple line) increased greatly. The increase corresponds to the importance of $y$ translation movement of the HCDR when the kinematic redundancy is resolved. Likewise, $ u_4 $ (solid green line) associated with the rotational motion around the $ x$-axis  reduced significantly.
	
	In the absence of the ILC, the UR3 manipulator link has entered the avoidance transition region at IT1 and IT2 (highlighted by yellow in Fig \ref{fig:spiderarm_simulation}), corresponding to the proximity of cable $ c_1 $ and $ c_2 $ (Fig \ref{fig:spiderarm_simulation} (a)), respectively. After the ILC implementation, the CDPR moved away from the trajectory along $ x $-axis since $ q_1 $ decreased at IT1 and IT2 (Fig \ref{fig:spiderarm_simulation} (i)). Similarly, rotation along the $ x $-, and $ z $-axis, given by $ q_4 $, and $ q_6 $, respectively (Fig \ref{fig:spiderarm_simulation} (a)), reduced significantly to prevent the manipulator from hitting the cables. The $ y $ movement of the HCDR is associated with the $ q_2 $ joint (Fig. \ref{fig:spiderarm_simulation} (a), purple color), which also increased drastically at IT1 and IT2 (Fig. \ref{fig:spiderarm_simulation} (i)). The increase in the translational $ y $ motion led to less required rotational movement of the UR3 $ q_7 $, $ q_8 $, and $ q_{11} $ joints (Fig. \ref{fig:spiderarm_simulation} (j)) to cover the same trajectory workspace, thereby preventing the UR3 link from hitting the cables $ c_1 $ and $ c_2 $ by not crossing below the cable-link buffer (Fig. \ref{fig:spiderarm_simulation} (f)). 
	
	\begin{figure}[t]
		\centering
		\includegraphics[width = \fss\columnwidth]{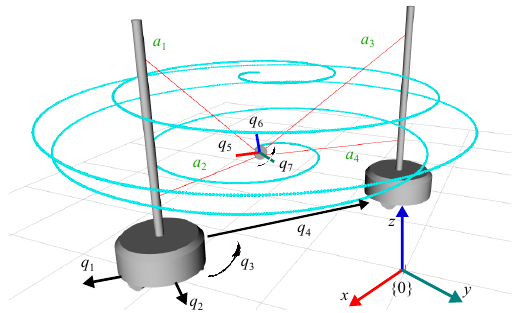}
		\caption{FASTKIT-Planar operational space tracking simulation results for spherical helix trajectory.}
		\label{fig:fastkit_simulation}
	\end{figure}
	
	In summary, the results show that the ILC was able to learn that for cable-link interference avoidance, the CDPR needs to move more laterally to reduce the UR3 joints motion. Hence, the results demonstrate that changes in $\btheta$ can affect the joint space motion of the robot and hence, it can improve task performance. Additionally, the operational space error and cable force norm also reduced after the implementation of the ILC (Fig \ref{fig:spiderarm_simulation} (k), and (l)).
	
	
	\subsection{FASTKIT-Planar} \label{sec:react_fastkit}
	
	The FASTKIT-Planar robot \citep{Rasheed_Long_Marquez_Caro_18_FASTKIT} consists of a planar CDPR between two mobile bases (Fig. \ref{fig:fastkit_simulation} (a)). The joint space is defined as $\bm{q} = [q_1, q_2, ..., q_7]^T$, where $q_1$ and $q_2$ refers to the \emph{xy}-coordinates of the first mobile base, and $q_3$ refers to its orientation. The distance between the two mobile bases is modelled by a prismatic joint, denoted as $q_4$. For the planar robot, $q_5$, $q_6$ and $q_7$ represent the \emph{xy}-coordinates and the orientation of the end-effector in the frame of the first mobile base. The planar robot is actuated by 4 cables $a_1, a_2, a_3$ and $a_4$, with 2 cables attached on each mobile base. Actuation forces created by each joint on the mobile bases are denoted as $a_5, a_6, a_7$ and $a_8$. The operational space is defined as $\bm{x} = [x_1, x_2, x_3]^T$, which represents the \emph{xyz}-coordinate of the end-effector of the planar robot in frame \{0\}. Next, the results obtained from the simulations performed on the FASTKIT-Planar HCDR are discussed.
	
	The design of the FASTKIT-Planar (Fig. \ref{fig:fastkit_simulation} (a)) adds two major challenges in the \ac{RC}. First, the mobile bases have unlimited workspace, and their movements can possibly have no effect on the end-effector. Second, the controller should keep the movement of the planar CDPR between the two mobile bases. Hence, for a system such as the FASTKIT-Planar with high number of redundant \ac{DoF}, the QP formulation (\ref{eqn:reactive_qp}) for the reactive controller allows the resolution of kinematic and actuation redundancies efficiently. 
	
	In the FASTKIT-Planar, the avoidance acceleration given by \eqref{eqn:avoid_fun_single} was formulated by combining the avoidance of low manipulability and cable-link interference acceleration as per \eqref{eqn:weight}. To prevent the planar robot from colliding with the mobile bases, the avoidance acceleration for cable-link interference was obtained by setting $h(\bm{q})$ to $\delta_{min}(\bm{q})$, where $\delta_{min}(\bm{q})$ refers to the minimium distance between the planar end-effector and the closest mobile base. The kinematic redundancy on the FASTKIT-Planar robot was explored and exploited with the use of the ILC on various trajectories. The FASTKIT-Planar simulations are discussed in Section \ref{sec:simulation_under_noise}.
	
	\begin{figure*}[!tbh]
		\centering
		\includegraphics[width=\fss\textwidth]{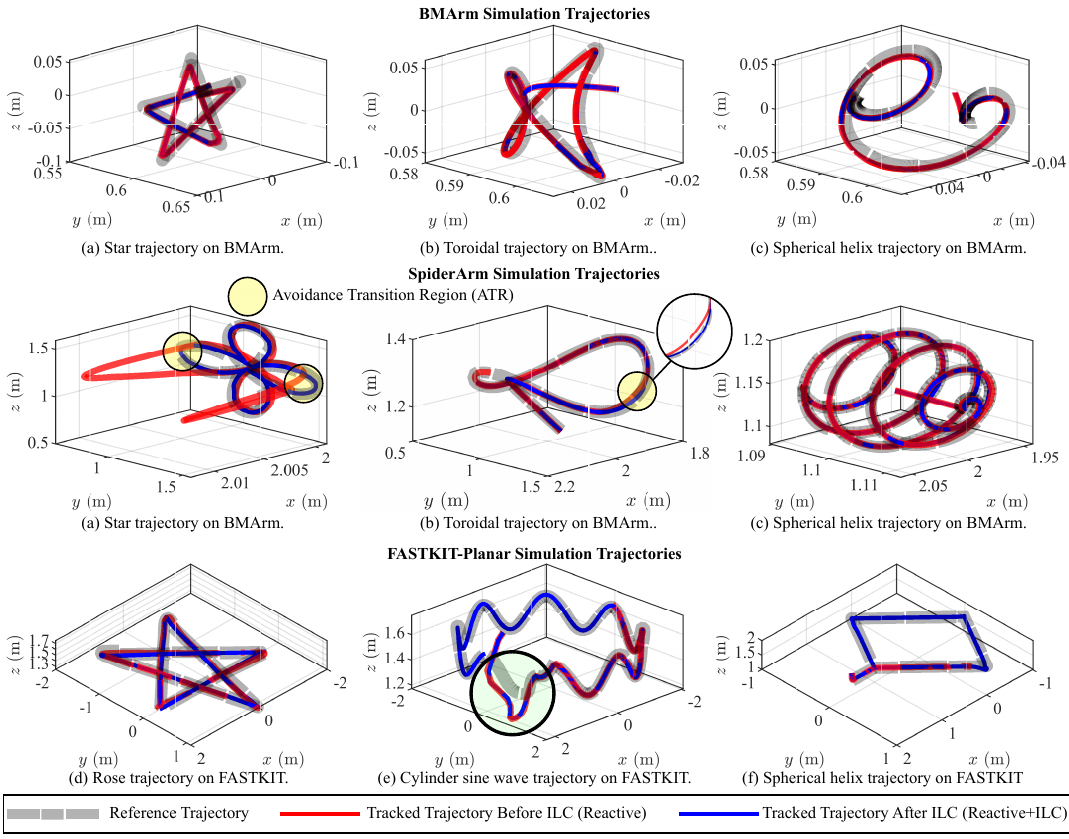}
		\caption{Simulation results of various trajectories with added noise to verify the the control framework robustness.}
		\vspace*{10mm}
		\label{fig:various_trajectories}
	\end{figure*}
	
	\begin{table*}[!tb]
		\centering
		\caption{Comparison of before and after ILC implementation impact on the Performance Indices (PIs) with noise.}
		\label{tab:perf_eval}
		\resizebox{0.90\textwidth}{!}{%
			\begin{tabular}{lcccccccccccccc}
				\toprule[0.8mm]
				\multicolumn{1}{c}{\bfseries CDPR}&\multicolumn{2}{c}{\textbf{\begin{tabular}[c]{@{}c@{}}Trajectory \\ with noise\end{tabular}}} & \multicolumn{6}{c}{\textbf{\begin{tabular}[c]{@{}c@{}}Root mean square\\ tracking error   (m)\end{tabular}}} & \multicolumn{2}{c}{\textbf{\begin{tabular}[c]{@{}c@{}}Average cable\\ force norm (N)\end{tabular}}}  & 
				\multicolumn{2}{c}{\textbf{\begin{tabular}[c]{@{}c@{}}Average direct\\ actuation norm ({N})\end{tabular}}} &
				\multicolumn{2}{c}{\textbf{\begin{tabular}[c]{@{}c@{}}Average \\ computation time (s)\end{tabular}}}\\ \midrule
				&\multicolumn{1}{c}{\multirow{2}{*}{\bfseries Name}}          &     \multirow{2}{*}{\bfseries Fig. \ref{fig:various_trajectories}}    &  \multicolumn{3}{c}{\textbf{Before ILC}}& \multicolumn{3}{c}{\textbf{After ILC}} &\multirow{2}{*}{\textbf{Before ILC}  }                             & \multirow{2}{*}{\textbf{After ILC}}   &\multirow{2}{*}{\textbf{Before ILC}} & \multirow{2}{*}{\textbf{After ILC}} & \multicolumn{1}{c}{\bfseries Controller} 
				& \multicolumn{1}{c}{\bfseries Net}      \\ \cline{4-9}
				& & &{$ x $}  & {$ y $}  & $ z $ & {$ x $}  & {$ y $}  & $ z $& & & & &      \\ \hline
				\multirow{3}{*}{BMArm}             &Star                   &      (a) & 9.87e-03 & 4.58e-04 & 1.16e-02 & 9.64e-03&4.49e-0   &1.16e-02  &45.6&45.0&-&-&    0.012   & 1.38      \\
				&Toroidal               &      (b) & 1.16e-03 & 1.85e-04 & 1.20e-03 & 7.77e-04 & 1.68e-04 & 1.20e-03 & 45.6 & 44.8 & - & - &  0.012   & 3.53  \\
				&Spherical helix        &      (c) & 4.30e-03 & 3.20e-04 &  5.12e-03 &  4.58e-03 & 3.15e-04 &  5.08e-03&43.5 & 43.4 & - & - & 0.012   & 3.37       \\ \midrule
				\multirow{3}{*}{SpiderArm}		   &Rose               & (d) & 3.57e-03 & 1.41e-02 & 9.03e-03 & 5.05e-04&1.92e-03 &1.96e-03&263&208&18.4&18.8&    0.014   & 5.83      \\ 
				&Cylinder sine wave & (e) & 1.90e-03 & 7.25e-03 & 2.05e-03 & 1.94e-03 & 7.73e-03 & 2.08e-03 & 265 & 192 & 18.6 & 15.4 & 0.006& 6.64      \\ 
				&Spherical helix    & (f) & 5.26e-03 & 5.41e-03 & 5.96e-03 & 0.80e-03 & 5.32e-03 & 0.93e-03 & 183 & 152 & 19.2 & 19.4 & 0.006& 2.29       \\ \midrule 
				\multirow{3}{*}{FASTKIT}		   &Star               & (g) & 3.40e-02 & 2.94e-02 & 4.59e-02 & 2.85e-02 &  1.36e-02 & 4.59e-02 & 23.5 & 21.6 & 126 & 91 & 0.006& 2.29      \\
				&Cylinder sine wave & (h) & Fail & Fail & Fail & 6.75e-02 & 4.44e-02 & 5.17e-02 & 265 & 37.7 & Fail & 43.8 & 0.007& 6.64      \\ 
				&Rectangular        & (i) & Fail & Fail & Fail & 6.68e-03 & 2.80e-03 & 3.99e-02 & Fail & 46.1 & Fail & 54.5 & 0.006& 2.29       \\ 
				\bottomrule[0.8mm]
			\end{tabular}%
		}
	\end{table*}
	
	\subsection{Robustness of the Proposed Framework with Various Trajectories}\label{sec:simulation_under_noise}
	
	In the earlier sections, detailed results were presented to demonstrate the behavior of the proposed framework. However, many more simulations were also conducted by adding white Gaussian noise to show the robustness of the proposed framework. Hence, in this section, a data set of many different trajectories were produced, and before and after ILC implementation simulations were run on them in presence of added noise to simulate the real-time sensor noise. For simulation comparison, the selected performance indicators are operational space Root Mean Square Error (RMSE) in tracking in $ x $, $ y $, and $ z $ direction, average cable force norm, and average direct actuation norm.  All the simulations simulation parameters were set as per the values set in Section \ref{sec:react_bmarm}, \ref{section:spiderarm_simulation_results}, and \ref{sec:react_fastkit}. The performance results are compared in Fig. \ref{fig:various_trajectories} and tabulated in Table \ref{tab:perf_eval}. Next we summarize the simulation results for the three CDPRs.

	\subsubsection{BMArm} Table \ref{tab:perf_eval} shows root mean square error and average cable force norm values for the BMArm. It showed that these values decreased slightly for all trajectories (Fig. \ref{fig:various_trajectories} (a) to (c)). For the toroidal trajectory ((Fig. \ref{fig:various_trajectories} (b)), the manipulability increased significantly from $ 3.37\times10^{-2} $ to $ 4.10\times10^{-1} $.
	
	\subsubsection{SpiderArm} From Table \ref{tab:perf_eval}, it was observed that the root mean square error for the SpiderArm tracking either decreased (Spherical helix) or remained almost the same (rose and cylindrical sine wave) before and after the ILC implementation. However, the ILC was able to improve the tracking error significantly at the IT (encircle and highlighted by yellow in Fig. (d) and (e)) by discovering a joint space trajectory, which would not allow the system to enter into the avoidance transition region. Additionally, the average cable force and direct actuation norms also decreased greatly. 
	\begin{figure*}[!h]
		\centering
		\includegraphics[width=\fss\textwidth]{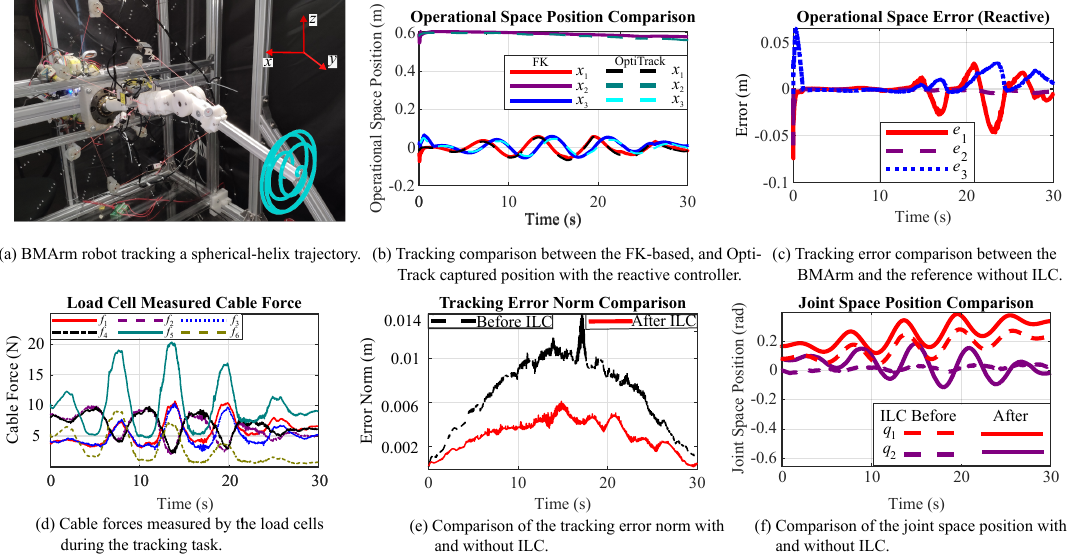}
		\caption{BMArm hardware operational space tracking results for spherical helix trajectory.}
		\label{fig:bmarm_realtime_reactive}
	\end{figure*}
	
	\subsubsection{FASTKIT-Planar} In the FASTKIT-Planar, after the ILC implementation, all the performance indicators decreased significantly (Fig (g)) for the star-shaped trajectory. However, with noise and without the ILC, the simulations shown in Fig. (h), and (i) failed for the FASTKIT-Planar, which became successful with the ILC. Additionally, Fig. \ref{fig:various_trajectories} (h) shows that in presence of random noise the tracking initially started poorly (encircled), but after $t=1.1$ s the system's end-effector movement resumed the reference trajectory path. For Fig. \ref{fig:various_trajectories} (h), and (i), while the reactive controller tracking failed at $ t=10 $ and $ t=5 $ s. However the ILC operated reactive controller was able to finish the tracking without divergence. 
	
	The analysis of the results show that the selection of $\alpha$ and $\beta$ are important for feasibility and performance of the controller framework.  By referring to \eqref{eqn:qp_obj},  a low value of $\alpha$ and $ \beta $ emphasized more on minimizing tracking error. For $\alpha=\beta=0$, a significant reduction in the root mean square error was found. However, for simulation with $\alpha = 100$, a greater reduction in the average cable force and direct actuation norms were observed. 

	\section{Hardware Results}\label{sec:hardware_results}
	
	The proposed control framework was applied on the BMArm robot hardware (Fig. \ref{fig:bmarm_realtime_reactive} (a)). The hardware was controlled by two computers: 1) a 64-bit computer, with a 3.4 GHz Core i7-6700 processor and 16 GB RAM, running the CASPR-ROS software platform \citep{Eden_Song_Tan_Oetomo_Lau_17_CASPRROS} responsible for the low-level motor control; and 2) with an Intel Core i9-11900K CPU @ 3.50 GHz and 32.0 GB of RAM, running CASPR in MATLAB for the proposed reactive and ILC framework.
	
	\subsection{Reactive Control}
	A helix-shaped trajectory, similar to that in Section \ref{sec:react_fastkit} was tracked, rotated about the \emph{y}-axis with a radius of $0.06$ m, thickness of $0.012$ m, and center at $[0,0.595,0]^T$. The starting joint pose was set at $\bm{q} = [0.08,0,0,-0.2]^T$. Control parameters were set at $\bm{K}_p=20000\bm{I}_3$, $\bm{K}_d=282\bm{I}_3$ (damping ratio $\approx 1$), $\alpha=1\times10^{-6}$, $\beta=1$. A high value was chosen for $\bm{K}_p$ to overcome the friction in the system caused by the routing of cables through the various pulleys.

	The BMArm's actual operational space position $ \bm{x}$ was obtained through \ac{FK} using the cable length feedback, which was calculated from the motor encoder feedback. To validate the FK determined position, Fig \ref{fig:bmarm_realtime_reactive} (b) compares the FK-based end-effector position with the position captured by the OptiTrack Prime 13 system. Fig. \ref{fig:bmarm_realtime_reactive} (c) shows the operational space tracking error between the hardware tracking task, and the reference trajectory in which an error amplitude lower than $0.05$ m was obtained. Cable forces measured by the load cells (Fig. \ref{fig:bmarm_realtime_reactive} (d)) also shows that the tracking task was achieved with positive cable forces. For the computational time of the reactive controller, a worst-case of $15.5$ ms was recorded during the warm-up stage of the QP process. However, $99$\% of the time steps achieved an average of $2.2$ ms, which demonstrates the capability of the proposed reactive controller to achieve tri-space operational control of CDPR hardware online in real-time. The obtained results are consistent with the BMArm simulations.

	\subsection{Iterative-Learning Control (ILC)}
	
	ILC was also performed on the BMArm hardware as per Section \ref{sec:ilc}. For optimizing \eqref{eqn:ilc_nonlin_op} and \eqref{eqn:ilc_P}, \ac{PSO} with $6$ particles was used to explore the solution space of $\btheta$ over $15$ iterations, resulting $6 \times 15 = 90$ evaluations. Position bounds for the particles were set at $\btheta_{min} = [-2,-2,-2,-2,\ln(1\times 10^{-7}),\ln(0.1)]^T$, $\btheta_{max} = [2,2,2,2,\ln(1\times 10^{-5}), \ln(10)]^T$. The inertia weight $\omega$ was set at $0.73$, while particle and swarm best parameters $\phi_p, \phi_g$ were both set at $1.5$ \citep{Kennedy_Eberhart_95_ParticleSwarmOptimization}.
	
	The results with and without ILC show that the tracking has improved, with the maximum amplitude of the error norm reducing from $0.014$ m to $0.006$ m (Figure \ref{fig:bmarm_realtime_reactive} (e)). Additionally, the cable forces used in the tracking task also reduced along the entire trajectory. Both improvements in the tracking error and cable force norm contributed a $ 64 \text{ }\% $ reduction in the performance function.
	
	The kinematic redundancy was resolved by the RC, which resulted in the joint space trajectory shown by the dashed lines in Fig. \ref{fig:bmarm_realtime_reactive} (a). Furthermore, by comparing the joint space trajectory before and after ILC implementation, it can be observed that the tracking error and cable force norm improvement is closely related to an increase in the movement of $q_2$, which corresponds to the twisting motion along the $y$-axis. Although excessive twisting creates potential risk for loss of manipulability (Section \ref{sec:avoid}), the ILC could determine a joint space trajectory that allowed twisting to a suitable extent, such that the tracking error and cable force needed for the task were reduced, while remaining stable. The average computational time recorded for updating $\btheta$ and evaluating $P(\btheta)$ after completing each trajectory are $0.75$ ms and $16.71$ ms, respectively, which imples that the proposed control framework could complete a new task and improve the task performance online on robot hardware.
	\acresetall
	\section{Conclusion}\label{sec:conclusion}
	This work proposed a novel tri-space operational control framework, dealing with actuation and kinematic redundancies, was proposed for a class of \ac{CDPR} including \ac{MCDR} and \ac{HCDR} when they are performing repetitive tasks. This framework consisted of both a \ac{RC}  and an \ac{ILC} systematically. The \ac{RC} was used for solving a convex \ac{QP} problem such that an operational space reference trajectory can be tracked with feasible cable forces, while avoiding undesirable situations. The ILC explored the best set of null space exploration parameters in the null space component of the joint space. It also tuned the objective function weights. Hence, the ILC component further improved the performance of the RC at a trajectory level. To demonstrate the capability and generalizability of the proposed framework, simulations were performed on one MCDR of BMArm, and two HCDRs, which are the SpiderArm and FASTKIT-Planar robot respectively. Experimental results on a two link MCDR of BMArm, were also presented to demonstrate the practicality of the framework in controlling hardware online. 

	\section*{Acknowledgement}
	The work was supported by the Research Grants Council (General Research Fund Reference No. 14203921) and the Innovation and Technology Commission (University-Industry Collaboration Programme Reference No. UIM/345).
	
	

	\begin{IEEEbiography}[{\includegraphics[width=1in,height=1.25in,clip,keepaspectratio]{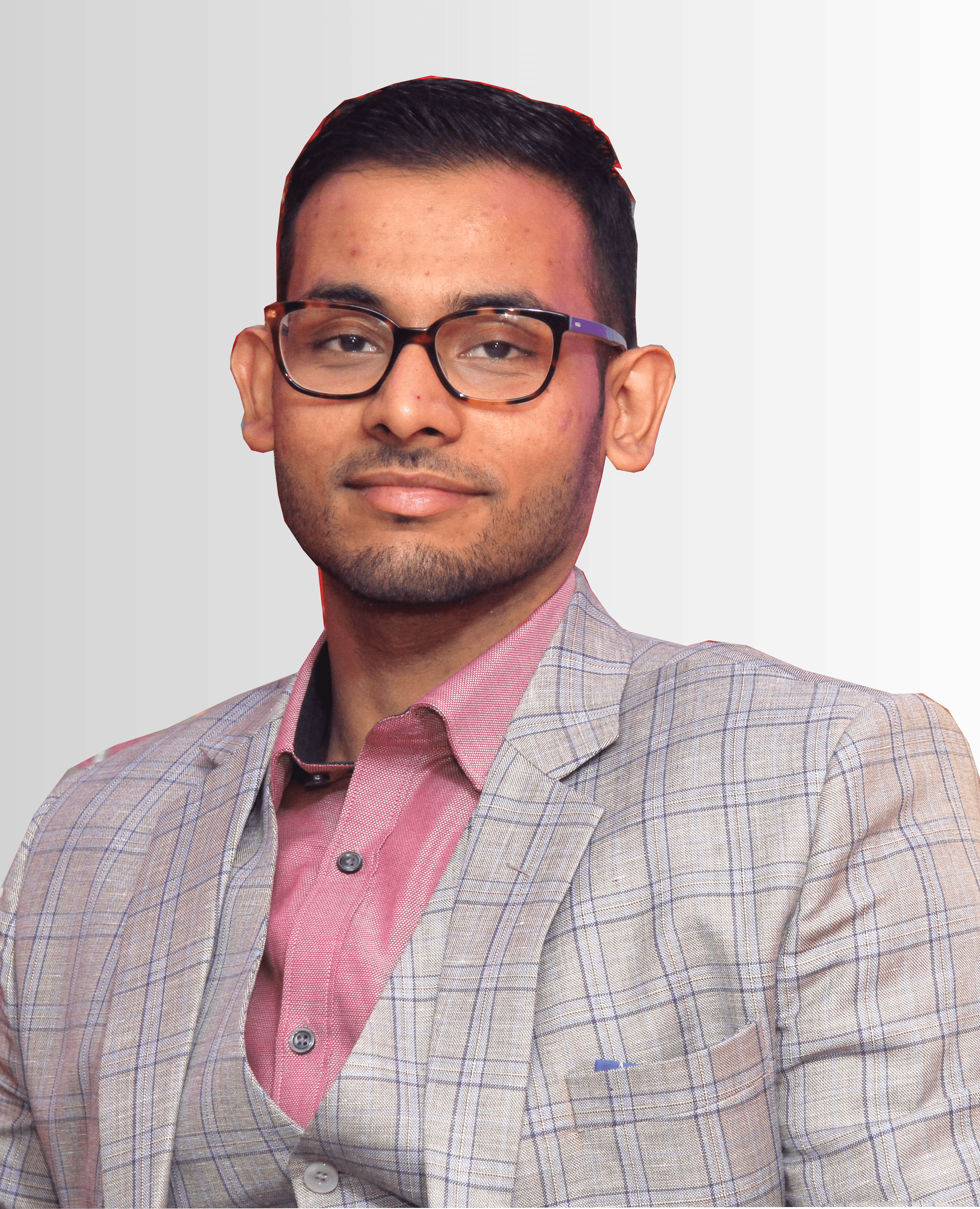}}]		{Dipankar Bhattacharya} Dipankar Bhattacharya (Member, IEEE) received the B.Tech.
		degree in electronics and communication engineering
		from the NERIST, India, in 2010, M.Tech. degree in electrical engineering from
		the IIT Roorkee, India, and the Ph.D. degree in mechatronics engineering from the University of Auckland, New Zealand in 2013, and 2021, respectively. He is currently working as a Postdoctoral Research Fellow in the Department of Mechanical and Automation engineering in the Chinese University of Hong Kong, Hong Kong. His research interests include biologically
		inspired soft robots, cable-driven parallel robots,  and robotics modeling and control.
	\end{IEEEbiography}

	\begin{IEEEbiography}[{\includegraphics[width=1in,height=1.25in,clip,keepaspectratio]{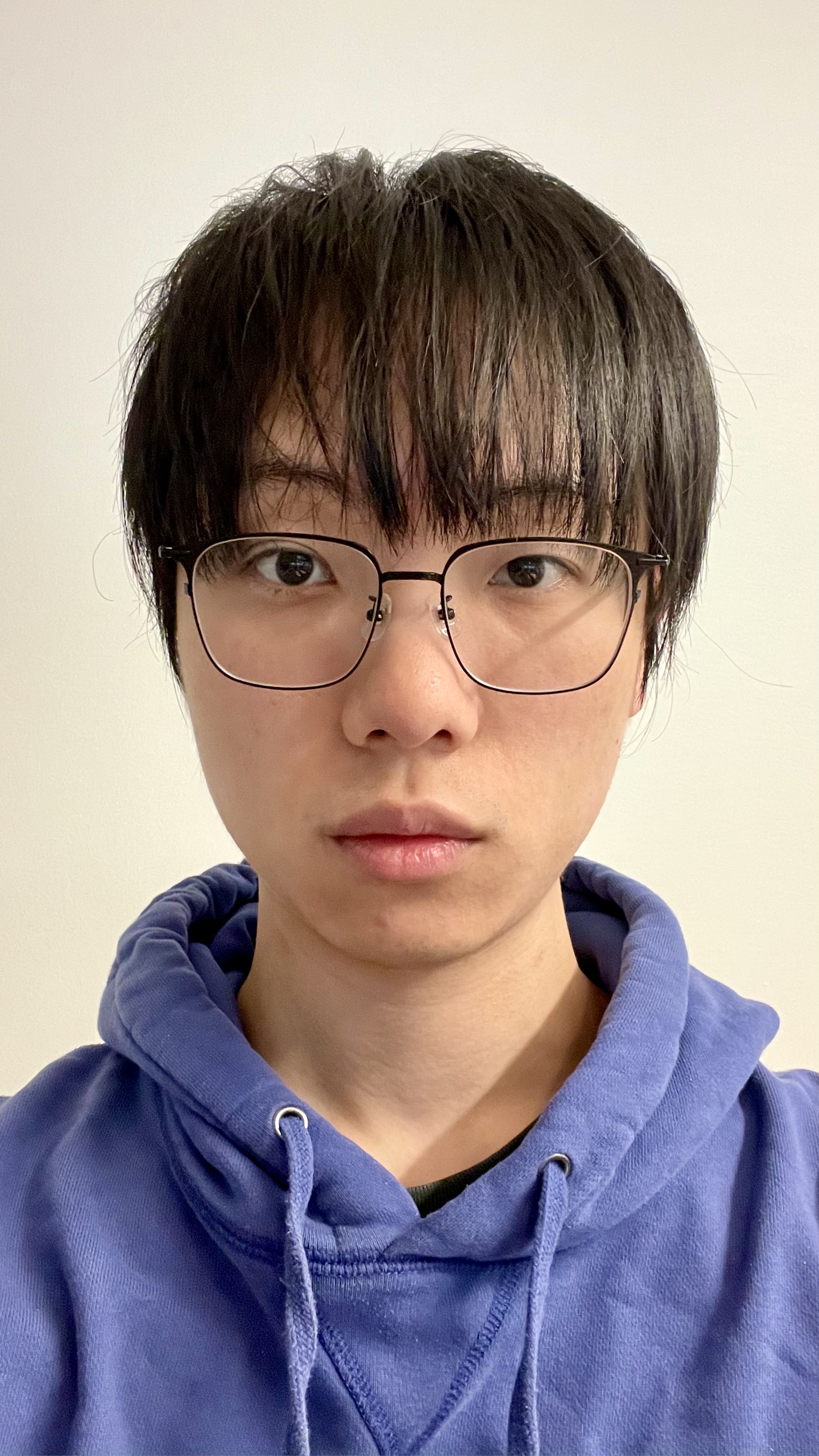}}]		{Yin Pok Chan}
	Yin Pok Chan received his B.Eng (2017) and M.Phil. (2019) degree in Mechanical and Automation Engineering from the Chinese University of Hong Kong. His research interest include cable robots, control systems, optimization and machine learning.
	\end{IEEEbiography}

	\begin{IEEEbiography}[{\includegraphics[width=1in,height=1.25in,clip,keepaspectratio]{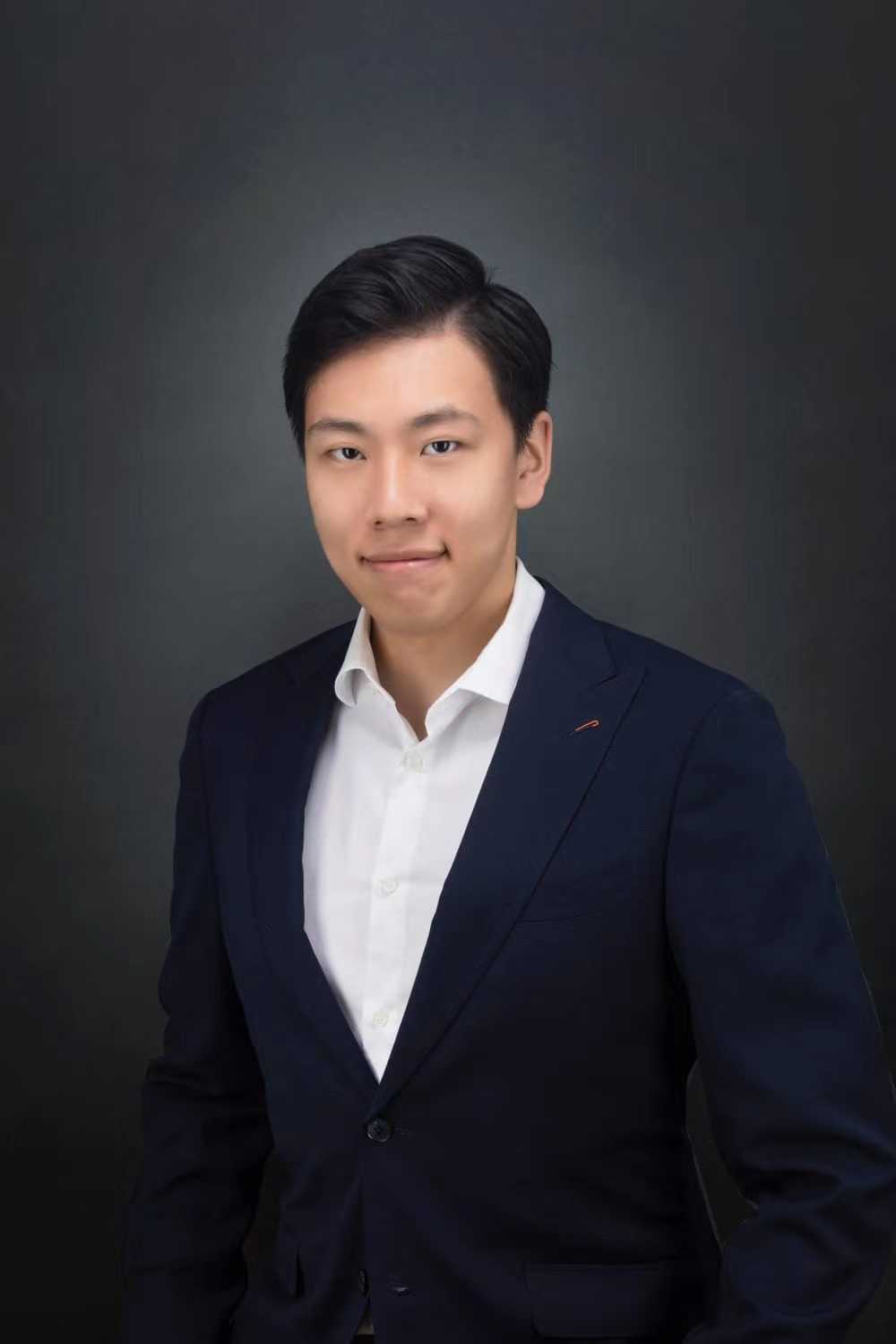}}]		{Siqi Shang}
	Siqi Shang received his Bachelors in Computer Science (2021) at the Chinese University of Hong Kong, Masters in Computer Science (2023) at Columbia University. His research interests include cable-driven robots, robotic learning, and dexterous manipulation.
	\end{IEEEbiography}

	\begin{IEEEbiography}[{\includegraphics[width=1in,height=1.25in,clip,keepaspectratio]{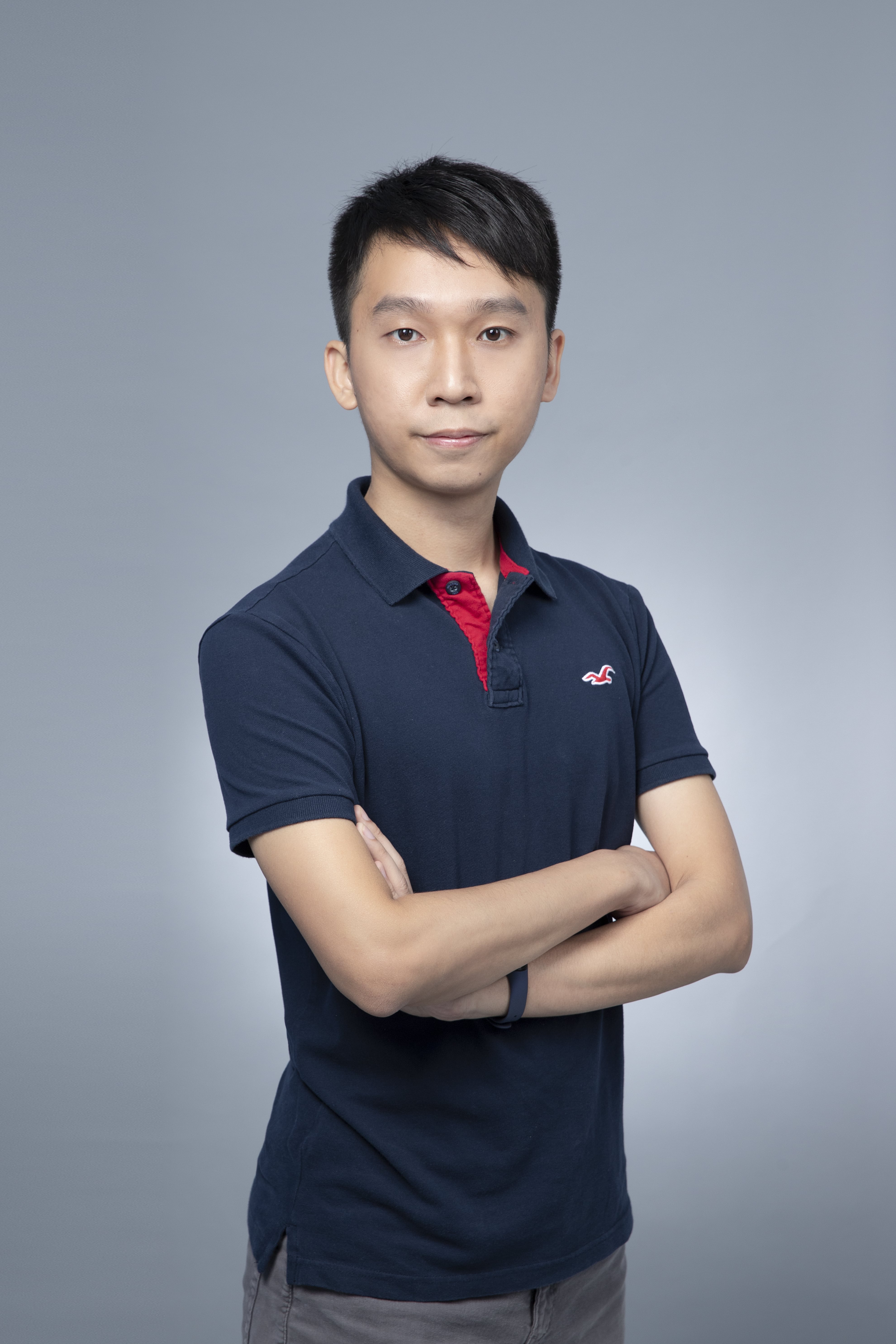}}]		{Yuen Chan Shan}
	Yuen Shan Chan received the B.Eng. and M.Phil. degrees mechanical and automation engineering from The Chinese University of Hong Kong in 2017 and 2019, respectively. For his M.Phil. studies, he specialized in the application of cable-driven robotics, particularly in large-scale maneuvering applications. In addition,
	Chan has also worked on various projects including underwater ROV and architecture perspective. 
	\end{IEEEbiography}

	\begin{IEEEbiography}[{\includegraphics[width=1in,height=1.25in,clip,keepaspectratio]{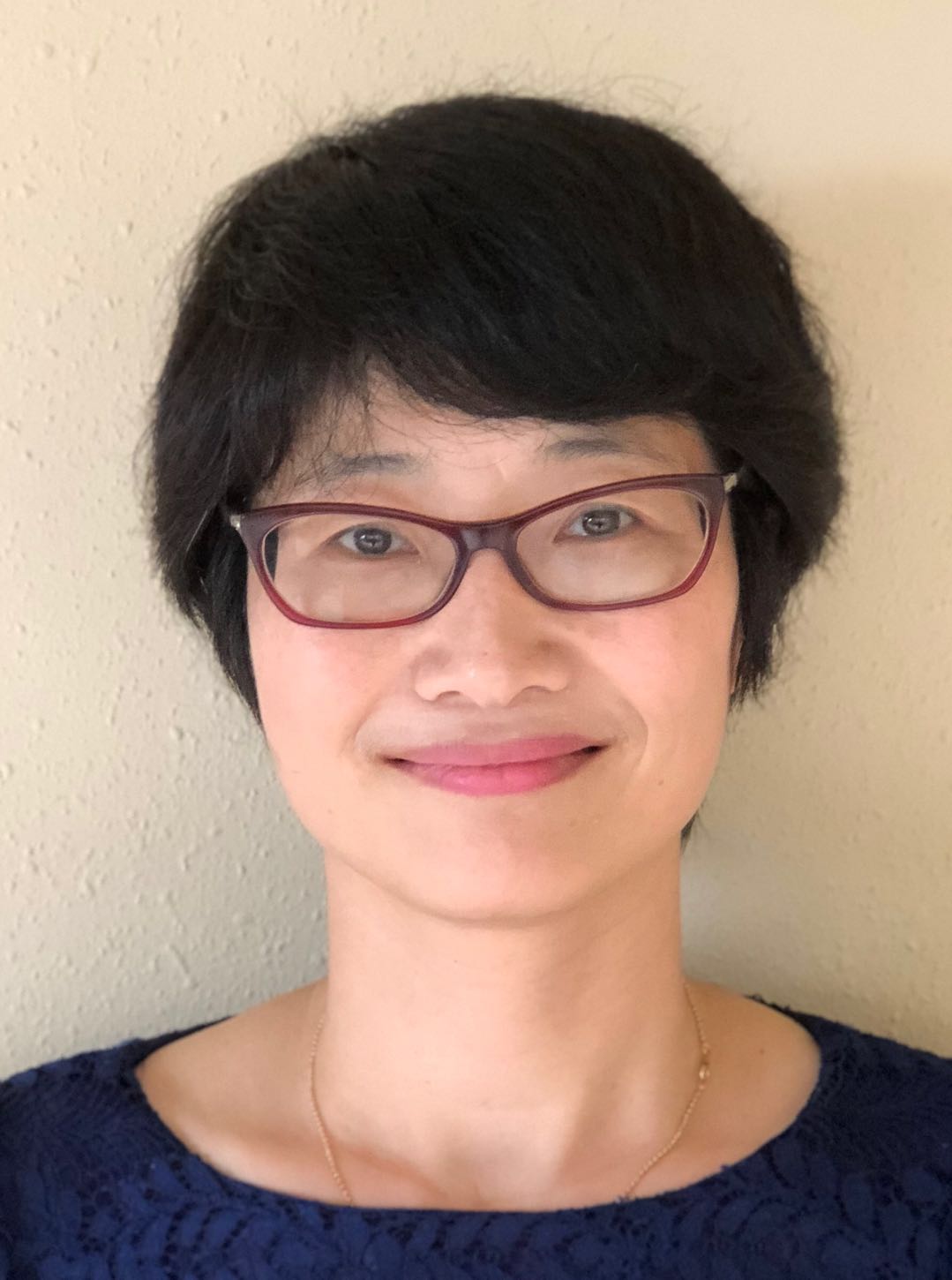}}]		{Ying Tan} 
	Ying Tan is a Professor in the Department of Mechanical Engineering
	at The University of Melbourne, Australia. She received her bachelor's
	degree from Tianjin University, China, in 1995, and her PhD from
	the National University of Singapore in 2002. She joined McMaster University in 2002 as
	a postdoctoral fellow in the Department of Chemical Engineering. Since 2004, she has been with the
	University of Melbourne. 
	She was awarded an Australian Postdoctoral Fellow (2006–2008) and a
	Future Fellow (2009–2013) by the Australian Research Council. She is Fellow of the Institute of Electrical and
	Electronic Engineers (FIEEE), Fellow of the Institution of Engineers of Australia (FIEAUST), and Fellow
	of Asia-Pacific Artificial Intelligence Association. Her research interests are in intelligent systems, nonlinear systems, real-time optimization, sampled-data systems, rehabilitation robotic systems, human motor learning, wearable sensors, and model-guided
	machine learning. 
	\end{IEEEbiography}

	\begin{IEEEbiography}[{\includegraphics[width=1in,height=1.25in,clip,keepaspectratio]{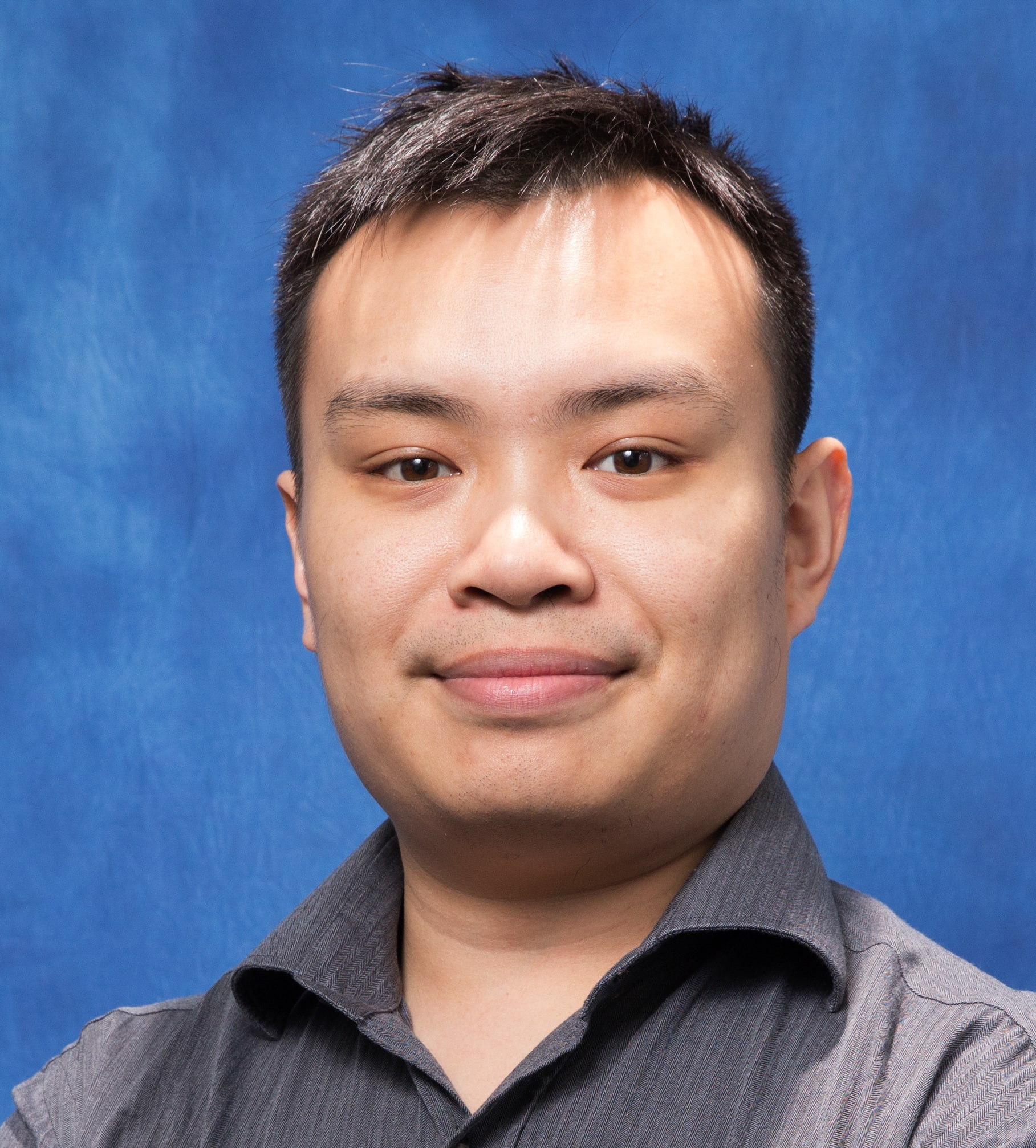}}]		{Darwin Lau} 
	Darwin Lau (Senior Member, IEEE) received the B.Eng. (Hons) degree in mechatronics engineering, B.CS. degree, and the Ph.D. degree in mechanical engineering and robotics from the University of Melbourne, Parkville, VIC, Australia, in 2008 and 2014, respectively. He was a Postdoctoral Research Fellow with Pierre and Marie Curie University from 2014 to 2015. He is currently an Associate Professor with the Department of Mechanical and Automation Engineering, Chinese University of Hong Kong, Hong Kong. His research interests include kinematics, dynamics and control of redundantly actuated mechanisms, cable-driven parallel manipulators, construction and architectural robotics, and bioinspired robots.
	\end{IEEEbiography}
	
\end{document}